\documentclass[letterpaper,twocolumn,10pt]{article}
\usepackage{usenix2019_v3}

\usepackage{tikz}
\usepackage{bm}
\usepackage{amsthm,amsmath,amsbsy}
\usepackage{amssymb}
\usepackage[ruled,linesnumbered,vlined]{algorithm2e}
\usepackage{booktabs}
\usepackage{multirow}
\usepackage{subcaption}
\usepackage{caption}
\usepackage{float}
\usepackage{pifont}
\usepackage{xcolor}
\usepackage{enumitem}
\usepackage{comment}
\setlist[itemize]{align=parleft,left=0pt..1em}
\usepackage{afterpage}% http://ctan.org/pkg/afterpage
\usepackage{makecell} %@yrchen: for making two rows in table
\usepackage{varwidth} %@yrchen: for splitting figure captions into two lines
\usepackage{stfloats}

\newlength{\oldtextfloatsep}

\newcommand{\RNum}[1]{\uppercase\expandafter{\romannumeral #1\relax}}

\newcommand{\sysname}{BGL}

\def\ie{\textit{i.e.}}
\newcommand{\mypara}[1]{\noindent{\bf {#1}}~}

\newcommand{\new}[1]{{\textcolor{black}{#1}}}
\SetCommentSty{mycommfont}

\usepackage{amsmath}
\DeclareMathOperator*{\argmax}{arg\,max}

\begin{document}

\title{\Large \bf BGL: GPU-Efficient GNN Training by Optimizing Graph Data I/O and Preprocessing}

\renewcommand{\thefootnote}{\fnsymbol{footnote}}
\author{
{\rm Tianfeng Liu$^{1,3}$\footnotemark[1],  Yangrui Chen$^{2,3}$\footnotemark[1],  Dan Li$^1$, Chuan Wu$^2$, Yibo Zhu$^3$, Jun He$^3$,}\\ {\rm Yanghua Peng$^3$, Hongzheng Chen$^{3,4}$, Hongzhi Chen$^{3}$, Chuanxiong Guo$^3$}\\
$^1$Tsinghua University, $^2$The University of Hong Kong, $^3$ByteDance, $^4$Cornell University
% copy the following lines to add more authors
% \and
% {\rm Name}\\
%Name Institution
} % end author



%\keywords{datasets, neural networks, gaze detection, text tagging}


%\input{abstract}

\maketitle

\footnotetext[1]{These authors have contributed equally to this work.}

\renewcommand{\thefootnote}{\arabic{footnote}}
%-------------------------------------------------------------------------------
\begin{abstract}
Graph neural networks (GNNs) have extended the success of deep neural networks (DNNs) to non-Euclidean graph data, achieving ground-breaking performance on various tasks such as node classification and graph property prediction. Nonetheless, existing systems are inefficient to train large graphs with billions of nodes and edges with GPUs. The main bottlenecks are  the process of preparing data for GPUs – subgraph sampling and feature retrieving. This paper proposes BGL, a distributed GNN training system designed to address the bottlenecks with a few key ideas.
First, we propose a dynamic cache engine to minimize feature retrieving traffic. By a co-design of caching policy and the order of sampling, we find a sweet spot of low overhead and high cache hit ratio.
Second, we improve the graph partition algorithm to reduce cross-partition communication during subgraph sampling.
Finally, careful resource isolation reduces contention between different data preprocessing stages. 
Extensive experiments on various GNN models and large graph datasets show that BGL significantly outperforms existing GNN training systems by 20.68x on average. %by up to 20.68x. %1.14x to 69x.

\end{abstract}
\section{Introduction}
\label{sec:intro}

% introduce GNN
Graphs, such as social networks~\cite{DBLP:conf/iclr/KipfW17, DBLP:conf/nips/HamiltonYL17}, molecular networks~\cite{DBLP:conf/nips/FoutBSB17}, knowledge graphs~\cite{DBLP:conf/ijcai/HamaguchiOSM17} and academic networks~\cite{DBLP:journals/qss/WangSHWDK20},  
provide a natural way to model a set of objects and their relationships.Recently, there is increasing interest in extending deep learning methods for graph data. 
Graph Neural Networks (GNNs)~\cite{DBLP:journals/debu/HamiltonYL17, DBLP:conf/iclr/KipfW17, DBLP:conf/iclr/VelickovicCCRLB18} have been proposed and shown to outperform traditional graph learning methods~\cite{zhou2018graph, DBLP:journals/corr/abs-1901-00596, DBLP:journals/corr/abs-1812-04202} in various applications such as node classification~\cite{DBLP:conf/iclr/KipfW17}, link prediction~\cite{DBLP:conf/nips/ZhangC18} and graph property prediction~\cite{DBLP:conf/aaai/YanXL18}. 
Real-world graphs can be huge. For example, the user-to-item graph at Pinterest contains over 2 billion entities and 17 billion edges with 18 TB data size~\cite{DBLP:conf/kdd/YingHCEHL18}. As a major online service provider, we also observe over 100 TB size of graph data, which consists of 2 billion nodes and 2 trillion edges.
Such large sizes make it impossible to load the entire graph into GPU memory (at tens of GB) or CPU memory (at hundreds of GB),
hence turning down proposals that adopt full graph training on GPUs~\cite{DBLP:journals/pvldb/ZhangHL0HSGWZ020}.
Recent works~\cite{DBLP:conf/nips/HamiltonYL17, DBLP:conf/kdd/YingHCEHL18, DBLP:conf/nips/Huang0RH18} have resorted to mini-batch sampling-based GNN training. Instead of loading the whole graph entirely, neighborhood information is aggregated on subgraphs. A web-scale graph is partitioned among distributed graph store servers over multiple machines. % and thus scales to large graphs. 
The training proceeds in iterations, each with three stages: (1) {\em sampling} subgraphs stored in distributed graph store servers, (2) {\em feature retrieving} for the subgraphs, and (3) forward and backward {\em computation} of the GNN model. 

The first two stages, which we refer to as {\em data I/O and preprocessing}, are often the performance bottlenecks in such sampling-based GNN training.
After analyzing popular GNN training frameworks ({\em e.g.,} DGL~\cite{ DBLP:journals/corr/abs-1909-01315}, PyG~\cite{DBLP:journals/corr/abs-1903-02428} and Euler~\cite{ali2019euler}), we made two key observations. 
(i) High data traffic for retrieving training samples: when the subgraph being sampled is stored across multiple graph store servers, 
there can be frequent cross-partition communication for sampling; retrieving corresponding features from the storage to worker machines also incurs large network transmission workload. (ii) Modern GPUs can perform the computation of state-of-the-art GNN models~\cite{DBLP:journals/debu/HamiltonYL17, DBLP:conf/iclr/KipfW17, DBLP:conf/iclr/VelickovicCCRLB18} quite fast, leading to high demand for data input. 
To mitigate this problem, Euler adopts
parallel 
feature retrieval; 
DGL and PyG prefetch the sampling results. Unfortunately, none of them fully resolves the I/O bottleneck. For example, we observe only around 10\% GPU utilization in a typical DGL training job on a large graph (\S\ref{sec:mov} and \S\ref{sec:eval}). It means around 90\% of GPU cycles are wasted.

In this paper, we propose {\sysname}, a GPU-efficient GNN training system for large graph learning, to accelerate training and thus achieve high GPU utilization (near 100\%).
Focusing on removing the data I/O and preprocessing bottleneck, we identify three 
key limitations of existing frameworks, namely, 1) too heavy network traffic of feature retrieving, 2) large cross-partition communication overhead during sampling due to naive graph partition algorithms, and 3) naive resource allocation leading to poor end-to-end performance. We address those challenges, respectively.

{\em First}, to minimize feature retrieving traffic, we co-design a dynamic cache policy and the sampling order of nodes. \mbox{PaGraph}~\cite{lin2020pagraph}, a state-of-the-art cache design for GNN training, explicitly avoids dynamic caching policy because of high overhead. However, we find that static cache (no replacement during training) has low hit ratios when the graphs are so large that only a small fraction of nodes can be cached. In contrast, we show that a FIFO policy has acceptable overhead and high hit ratios once combined with our {\em proximity-aware ordering}. The key idea is to leverage {\em temporal locality} -- in nearby mini-batches, we always try to sample nearby nodes. It largely increases the cache hit ratio of FIFO policy. We will further explain the details of how we ensure the consistency of our multi-GPU cache engine and GNN convergence in \S\ref{sec:feature_cache_engine}.

{\em Second}, to minimize cross-partition communication during sampling, we design a %simple but effective 
graph partition algorithm tailored for the typical sampling algorithms in GNN training.
Existing partition algorithms either do not scale to large graphs or fail to consider {\em multi-hop
neighbor} connectivity inside each partition. It leads to heavy cross-partition communication because, in GNN training, the sampling algorithm usually requests {\em multi-hop neighbors} from a given node.
In contrast, our partition algorithm (in \S\ref{sec:design_partition_assignment}) strives to maintain multi-hop connectivity within each partition, meanwhile load balance partitions and is scalable to giant graphs.

{\em Finally}, we optimize the resource allocation of data preprocessing by profiling-based resource isolation. Data preprocessing in GNN consists of multiple operations that may compete for CPU and bandwidth resources. However, existing frameworks largely ignore it and let the preprocessing operations freely compete with each other. Unfortunately, some operations do not scale well with more resources but may acquire more resources than it actually needs, leading to low end-to-end preprocessing performance. Our key idea is to formulate the resource allocation problem as an optimization problem, use profiling to find out the resource demands of each operation, and isolate the CPU cores for each operation.

We implement {\sysname}, including all the above design points, and replace the data I/O and preprocessing part of DGL with it. The design of {\sysname} is generic -- for example, it can also be used with Euler's computation backend. However, our evaluation focuses on using {\sysname} with the DGL GPU backend because it is more mature and performant.
We conduct extensive experiments using multiple representative GNN models with various graph datasets, including the largest publicly available  dataset and an internal billion-node dataset.
We demonstrate that {\sysname} outperforms existing frameworks and the geometric mean of speedups over PaGraph, PyG, DGL and Euler is 2.14x, 3.02x, 7.04x and 20.68x, respectively.
With the same GPU backend as DGL, {\sysname} can push the V100 GPU utilization to 99\% even when graphs are stored remotely and distributedly, higher than existing frameworks by far.
It also scales well with the size of graphs and the number of GPUs.

\section{Background and Motivation}
\label{sec:mov}

\subsection{Sampling-based GNN Training}

We start by explaining sampling-based GNN training.

\mypara{Graph.} 
The most popular GNN tasks are to train on graphs with node features, $\mathcal{G} = (\mathcal{V},  \mathcal{E},\mathcal{F})$, where $\mathcal{V}$ and $\mathcal{E}$ respectively denote the node set and edge set of the graph, and
$\mathcal{F}$ denotes the set of feature vectors assigned to each node.
For example, in the graph dataset Ogbn-papers~\cite{DBLP:journals/qss/WangSHWDK20}, each node  (\ie, paper) has a 128-dimensional feature vector representing the embeddings of the paper title and abstract.

\mypara{Graph neural networks (GNNs).}
Graph neural networks are  
neural networks learned from graphs.
The basic idea is collectively aggregating information following the graph structure and performing various feature transformations.
For instance, the Graph Convolution Network (GCN)~\cite{DBLP:conf/iclr/KipfW17} generalizes the convolution operation to graphs. 
For each node, GCN aggregates the features of its neighbors using a weighted average function and feeds the result into a neural network.
For another example, GraphSAGE~\cite{DBLP:conf/nips/HamiltonYL17} is a graph learning model that uses neighbor sampling to learn different aggregation functions on different numbers of hops. 

Real-world graphs, such as social networks and e-commerce networks~\cite{DBLP:journals/pvldb/ZhangHL0HSGWZ020, DBLP:conf/kdd/YingHCEHL18, DBLP:journals/pvldb/ChingEKLM15}, are often large. The Pinterest graph~\cite{DBLP:conf/kdd/YingHCEHL18} consists of 2B nodes and 17B edges, and requires at least 18 TB memory during training. Even performing simple operations for all nodes would require significant computation power, not to mention the notoriously computation-intensive neural networks. Similar to other DNN training tasks, it is appealing to use GPUs to accelerate GNN training.

\mypara{Sampling-based GNN training.}
There are two camps of training algorithms adopted in existing GNN systems: \emph{full-batch training} and \emph{mini-batch training}.
Full-batch training loads the entire graph into GPUs for training%is first adopted for learning GCN models
~\cite{DBLP:conf/iclr/KipfW17}, like NeuGraph~\cite{DBLP:conf/usenix/MaYMXWZD19} and ROC~\cite{DBLP:conf/mlsys/JiaLGZA20}.
Unfortunately, for very large graphs like Pinterst's, such an approach would face the limitation of GPU memory capacity. 

Thus, we focus on the other approach, {\em mini-batch training}, or often called {\em Sampling-based GNN training}. In each iteration, this approach samples a subgraph from the large original graph to construct a mini-batch as the input to neural networks. 
Mini-batch training is more popular and adopted by literature~\cite{DBLP:conf/nips/HamiltonYL17,DBLP:conf/iclr/ChenMX18,DBLP:conf/iclr/ZengZSKP20} and popular GNN training frameworks like DGL~\cite{DBLP:journals/corr/abs-1909-01315}, PyG~\cite{fey2019fast} and Euler~\cite{ali2019euler}.

%%!!! replace figure this figure by following pesudocodes
\begin{figure}[t]
\centering
\includegraphics[width=0.8\linewidth]{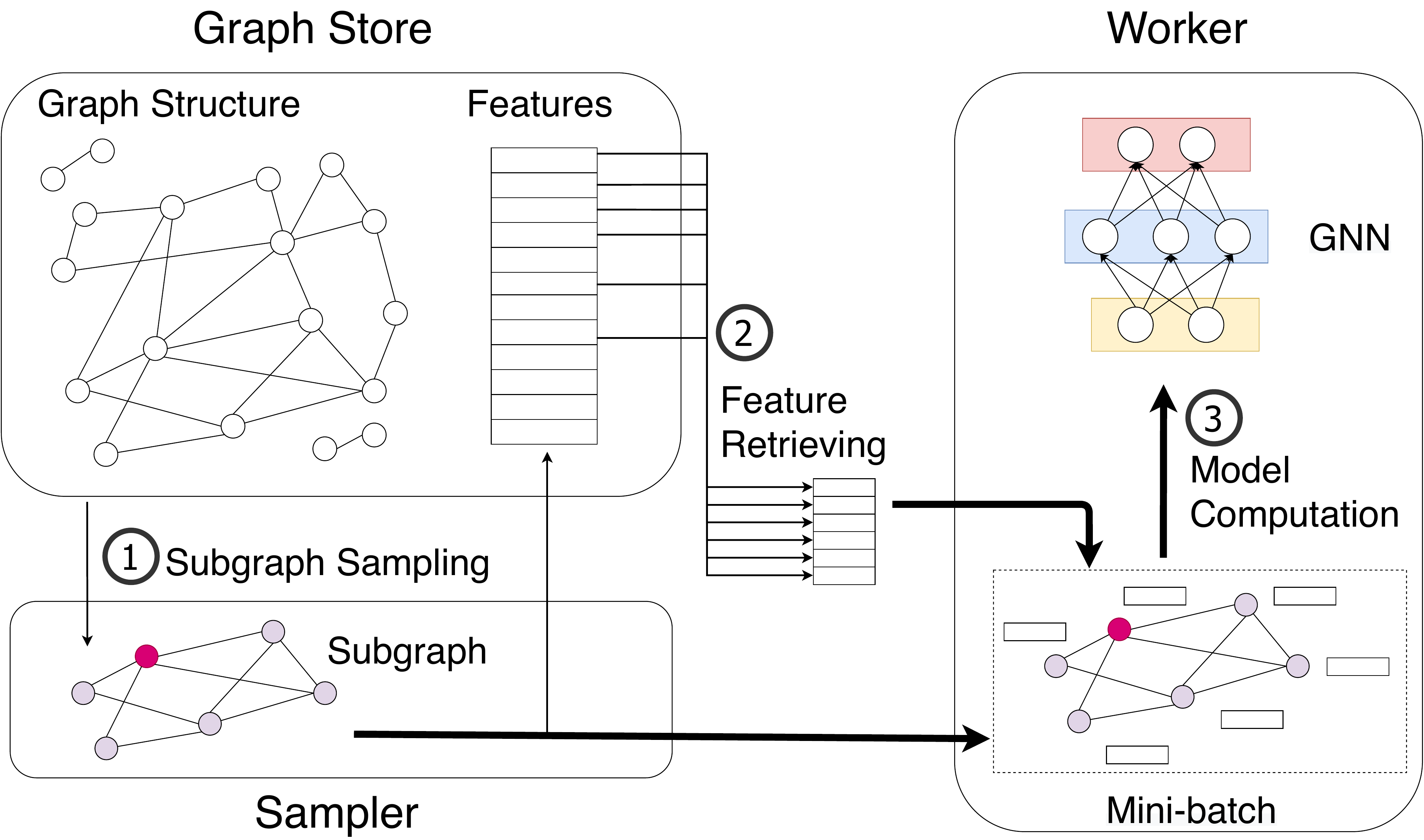}
\caption{Sampling-based GNN training process.}
\label{fig:mov_sampling}
\vspace{-5mm}
\end{figure}

The process of sampling-based GNN training is shown in Figure~\ref{fig:mov_sampling}.
The graph data (including the graph structure and node features) are partitioned and stored in a distributed \textit{graph store}.
Each training iteration consists of three stages:

\emph{(1) Subgraph sampling}. 
\textit{Samplers} sample a subgraph from the original graph and send it to workers.

\emph{(2) Feature retrieving}.
After \emph{workers} receive the subgraph,
the features of its nodes are further retrieved from the graph store server and placed in GPU memory.

\emph{(3) Model computation}. 
Like typical DNN training, workers forward-propagate the prepared mini-batch through the GNN model, calculate the loss function and then compute gradients in backward propagation. 
Then model parameters are updated using optimizers ({\em e.g.,}  SGD~\cite{DBLP:conf/nips/ZinkevichWSL10}, Adam~\cite{DBLP:journals/corr/KingmaB14}).

In the rest of this paper, we refer to the first two stages as {\em Data I/O and Preprocessing}.

\subsection{Data I/O and Preprocessing Bottlenecks}

Unfortunately, existing GNN training frameworks suffer from data I/O and preprocessing bottlenecks, especially when running model computation on GPUs. 
Here, we test two representative frameworks, DGL~\cite{DBLP:journals/corr/abs-1909-01315} and Euler~\cite{ali2019euler}. 
We train GraphSAGE~\cite{DBLP:conf/nips/HamiltonYL17} model with one GPU worker.
Using the partition algorithms of DGL and Euler, we split Ogbn-products graph~\cite{DBLP:journals/corr/abs-2005-00687} into two partitions and store them on two servers as a distributed graph store. More configuration details and the other framework results are in \S\ref{sec:eval}.

\begin{figure}[t]
\centering
\begin{minipage}[t]{0.48\linewidth}
\centering

\includegraphics[width=0.9\linewidth]{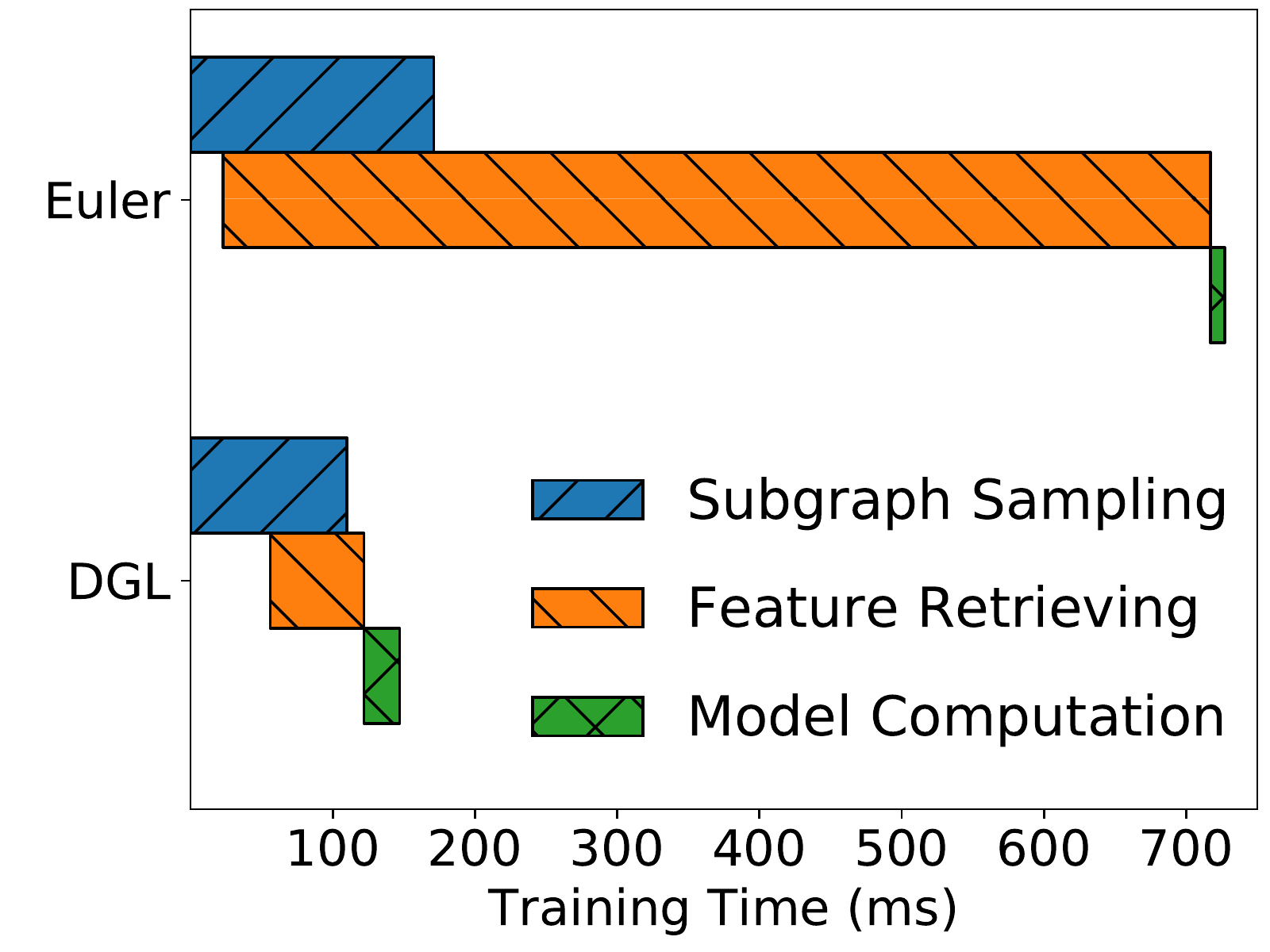}
%\vspace{-2mm}
\caption{Training time per mini-batch of DGL and Euler.} 
\label{fig:training_time}
\end{minipage}
\hspace{1mm}
\begin{minipage}[t]{0.48\linewidth}
\centering
\includegraphics[width=\linewidth]{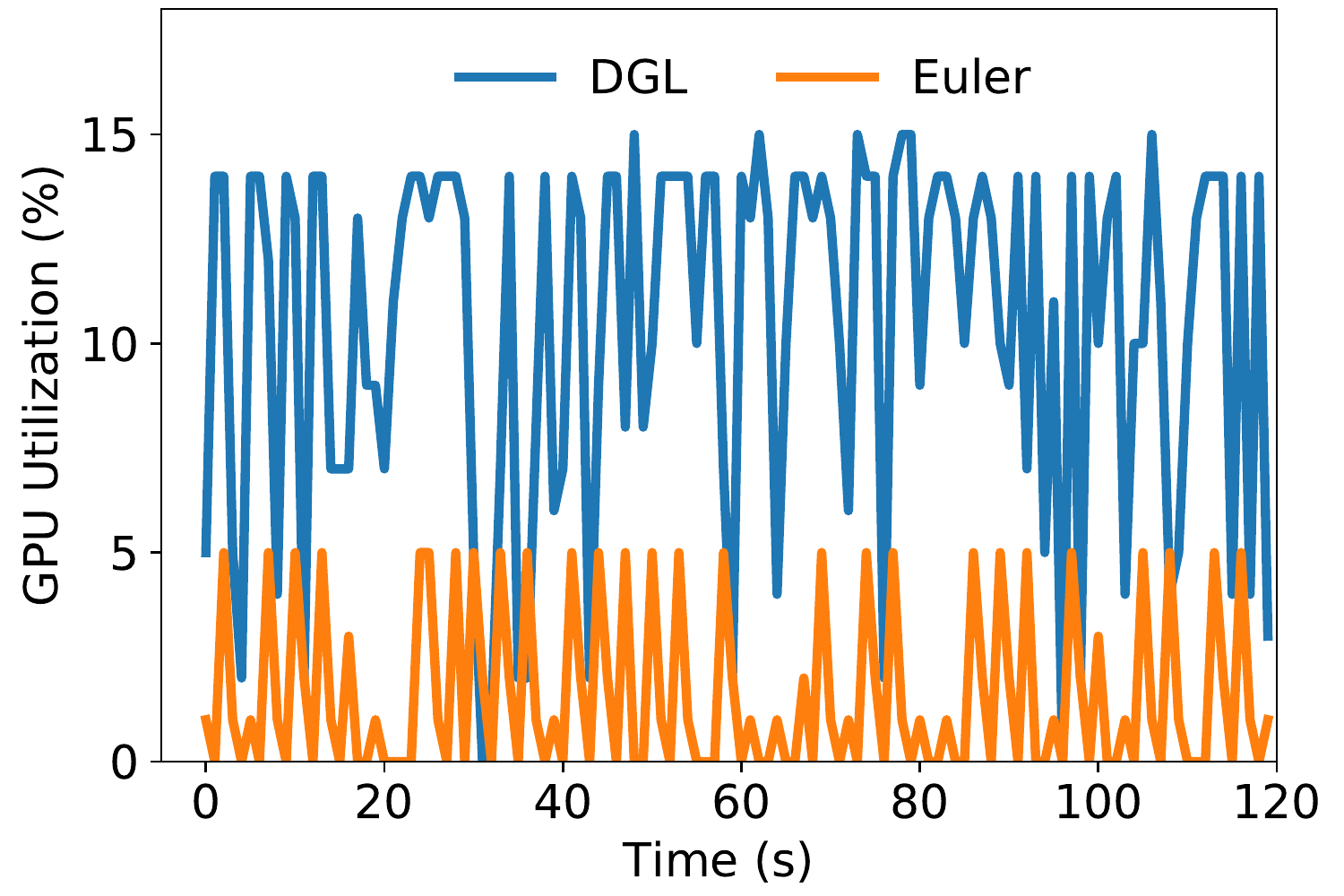}
\caption{GPU utilization of DGL and Euler.}
\label{fig:gpu_util_dgl}
\end{minipage}

\vspace{-5mm}
\end{figure}

Figure~\ref{fig:training_time} shows the training time of one mini-batch and the time breakdown of each stage.
99\% and 83\% of the training time were spent in data I/O and preprocessing by Euler and DGL, respectively.
Long data preprocessing time leads to not only poor training performance but also low GPU utilization. 
The maximum GPU utilization of DGL and Euler is 15\% and 5\%, respectively, as shown in Figure~\ref{fig:gpu_util_dgl}.

In GNN training, such bottleneck is much more severe than in DNN training like computer vision (CV) or natural language processing (NLP) for two main reasons.

First, due to the neighbor explosion problem~\cite{DBLP:conf/iclr/ZengZSKP20,DBLP:conf/kdd/ChiangLSLBH19}, the size of mini-batch data required by each training iteration is very large.
For example, if we sample a three-hop subgraph from Ogbn-products with batch size 1,000 and fan out \{15,10,5\}, each mini-batch consists of 5MB subgraph structure (roughly 400,000 nodes) and 195 MB node features. 
Assuming that we use a common training GPU server like AWS p3dn.24xlarge~\cite{amazonec2} (8x NVIDIA V100 GPUs and 100Gbps NIC) as the worker, and that we could saturate the 100Gbps NIC pulling such data, every second we can only pull 60 mini-batches of data. In addition, other preprocessing stages can further slow down the mini-batch preparation to as low as a few mini-batches per second (Figure~\ref{fig:training_time}).

Second, the model sizes and required FLOPS of GNN are much smaller than classic DNN models like BERT~\cite{devlin2018bert} or ResNet~\cite{he2016deep}. V100 needs only 20ms to compute a mini-batch of popular GNN models like GraphSAGE. 8 GPUs on p3dn.24xlarge can compute 400 mini-batches per second.

There is clearly a huge gap between the data I/O and preprocessing speed and GPU computation speed. Consequently, though frameworks like DGL and Euler adopt pipelining, the data I/O and preprocessing bottlenecks can only be hidden by a small fraction and dominate the end-to-end training speed.

Some recent work~\cite{lin2020pagraph,gandhi2021p3,DBLP:conf/eurosys/JangdaPGS21} also observed this problem and made promising progress. Unfortunately, it still falls short in performance (\S\ref{sec:eval}) and cannot handle giant graphs well. Next, we will elaborate the main challenges faced by existing GNN training frameworks.

\subsection{Challenges in Removing the Bottlenecks}
\label{sec:causes_and_opp}

We identify three main challenges that have not been fully resolved or even ignored by existing works.

\mypara{Challenge 1: Minimize the overhead of node feature retrieving.}
In the experiments shown in Figure~\ref{fig:training_time}, node feature retrieving may become the bottleneck due to the large volume of data being pulled to workers. A natural idea to minimize such overhead is to leverage the power-law degree distribution~\cite{faloutsos1999power} of real-life graphs. For example, PaGraph~\cite{lin2020pagraph} adopted a static (no replacement at runtime) cache that locally stores the predicted hottest node features. Upon cache hit, the traffic of feature retrieving can be saved. Unfortunately, on giants graphs like Pinterest graph~\cite{DBLP:conf/kdd/YingHCEHL18}, such static cache may only be able to store a small fraction of nodes due to memory constraint. In our experiments, when only 10\% of nodes can be cached, static cache only yields <40\% cache hit ratios.

Why not use dynamic (replacing some caches at runtime) cache policies? It is challenging because it would incur large searching and updating overhead, pointed out in~\cite{lin2020pagraph}. Overheads become even larger when cache is large (tens of GB) and stored on GPU. Our experimental implementation echos~\cite{lin2020pagraph} -- we also find that popular policies like LRU and LFU lead to near 80-millisecond overhead for updating. 

Nevertheless, we will show in \S\ref{sec:feature_cache_engine} that it is still possible to achieve a good trade-off between cache hit ratios and dynamic cache overhead, by exploiting the characteristics of GNN training and carefully designing the cache engine.

\begin{table}[!t]
    \centering
        \begin{minipage}[]{\columnwidth}
        \setlength{\tabcolsep}{4pt}
        \centering
        \caption{Qualitative comparison of different graph partition algorithms.}
        \label{tab:mov_partition}
        \begin{small}
        \begin{tabular}{c|c|c|c}
        \toprule
        {\footnotesize \textbf{\makecell{Partition \\ Algorithms}}} & {\footnotesize\textbf{\makecell{Scalability to \\ Giant Graphs}}} & {\footnotesize\textbf{\makecell{Balanced  \\Training Nodes}}} & {\footnotesize\textbf{\makecell{Multi-hop \\Connectivity}}} \\
        \midrule
        Random~\cite{jia2017distributed,ali2019euler}  & \ding{51} & \ding{51}& \ding{55}  \\
        \midrule
        \makecell{METIS\cite{george1998metis} \& \\ ParMETIS\cite{DBLP:journals/jpdc/KarypisK98}} & \ding{55}  & \ding{51} & \ding{51} \\
        \midrule
        GMiner~\cite{DBLP:conf/eurosys/ChenLZYYC18}  & \ding{51}  &  \ding{55} & \ding{55} \\
        \midrule
        PaGraph~\cite{lin2020pagraph} &\ding{55}& \ding{51} & \ding{51}\\
        \bottomrule
        \end{tabular}
        \end{small}
    \end{minipage}
\vspace{-5mm}
\end{table}

\mypara{Challenge 2: Develop a scalable partition algorithm for efficient sampling during GNN training.}
Giant graphs must be partitioned because a single server cannot hold the entire graph in its DRAM. Such algorithms must be scalable to giant graphs. 
Some partition algorithms, like the METIS~\cite{george1998metis} family used by DGL, rely on maximal matching to coarsen the graph (\S\ref{sec:graph_partition_module}), which is not friendly to giant graphs due to high memory complexity~\cite{DBLP:journals/pvldb/HanaiSTLTC19}.
Some other partition algorithms have high time complexity, like the one used by PaGraph~\cite{lin2020pagraph}, and are not friendly to giant graphs, either.

The partition algorithms are important because they affect the sampling overhead in two ways. First, they determine cross-partition communication overhead. Sampling algorithms construct a subgraph by sampling from a training node's {\em multi-hop} neighbors. If the neighbors are hosted on the same graph store server, {\em sampler} processes colocated with graph store servers can finish sampling locally. Otherwise, it must request data from other servers, incurring high communication overhead. Naive algorithms, like random partitioning\footnote{Also including Lux~\cite{jia2017distributed}, which is a random partition algorithm that frequently re-partitions the graph for load balancing.} used by Euler, are agnostic to the graph structure. 
Most state-of-the-art (SOTA) graph partition algorithms on graph processing and graph mining, like GMiner~\cite{DBLP:conf/eurosys/ChenLZYYC18} and CuSP~\cite{DBLP:conf/ipps/HoangDGP19}, only consider one-hop connectivity instead of multi-hop connectivity. They are suboptimal because  multi-hop neighbors are likely scattered in different graph partitions.

Second, partition algorithms determine the load balance across graph store servers and sampler processes. In a training epoch, one must iterate all {\em training nodes} and sample subgraphs based on them. For good load balance, one should balance the training nodes across partitions. 
However, SOTA graph partition algorithms only consider balancing all the nodes, of which only 10\%~\cite{DBLP:journals/corr/abs-2005-00687,DBLP:journals/qss/WangSHWDK20} are training nodes. 
Because they focus on maintaining neighborhood connectivity, they may produce less balanced partitions than the pure random algorithm, especially imbalanced for the training nodes.

Ideally, we need a partition algorithm that works on giant graphs and simultaneously minimizes the cross-partition communication and load imbalance during sampling. 
As shown in Table~\ref{tab:mov_partition}, none of the existing partition algorithms satisfies our needs, which motivates our algorithm (\S\ref{sec:graph_partition_module}).

\mypara{Challenge 3: Address the resource contention among multiple data preprocessing stages.}
We further identify a unique problem of GNN training -- the preprocessing is much more complex than
traditional DNN training. The subgraph sampling process, subgraph structure serialization and de-serialization, node feature retrieving and cache engine all consume CPU and memory/PCIe/network bandwidth resources. We find that, if all the processes freely compete for resources, the resource contention may lead to poor performance.
A key reason is that some operations do not scale well with more resources, while they may acquire more than they need for not becoming the bottleneck. 

Existing GNN training frameworks largely ignore this problem. DGL, PyG and Euler either blindly let all processes freely compete, or leave the scheduling to underlying frameworks like TensorFlow and PyTorch. The low-level frameworks are agnostic to the specifics in GNN training, thus are also naive and suboptimal. Our answer to this challenge is a carefully designed resource isolation scheme (\S\ref{sec:resource_isolation}).

\section{Design}
\label{sec:design}

\subsection{Overview}
\label{subsec:arch_overview}

\begin{figure}[t]
\begin{small}
 \centering
\includegraphics[width=0.9\linewidth]{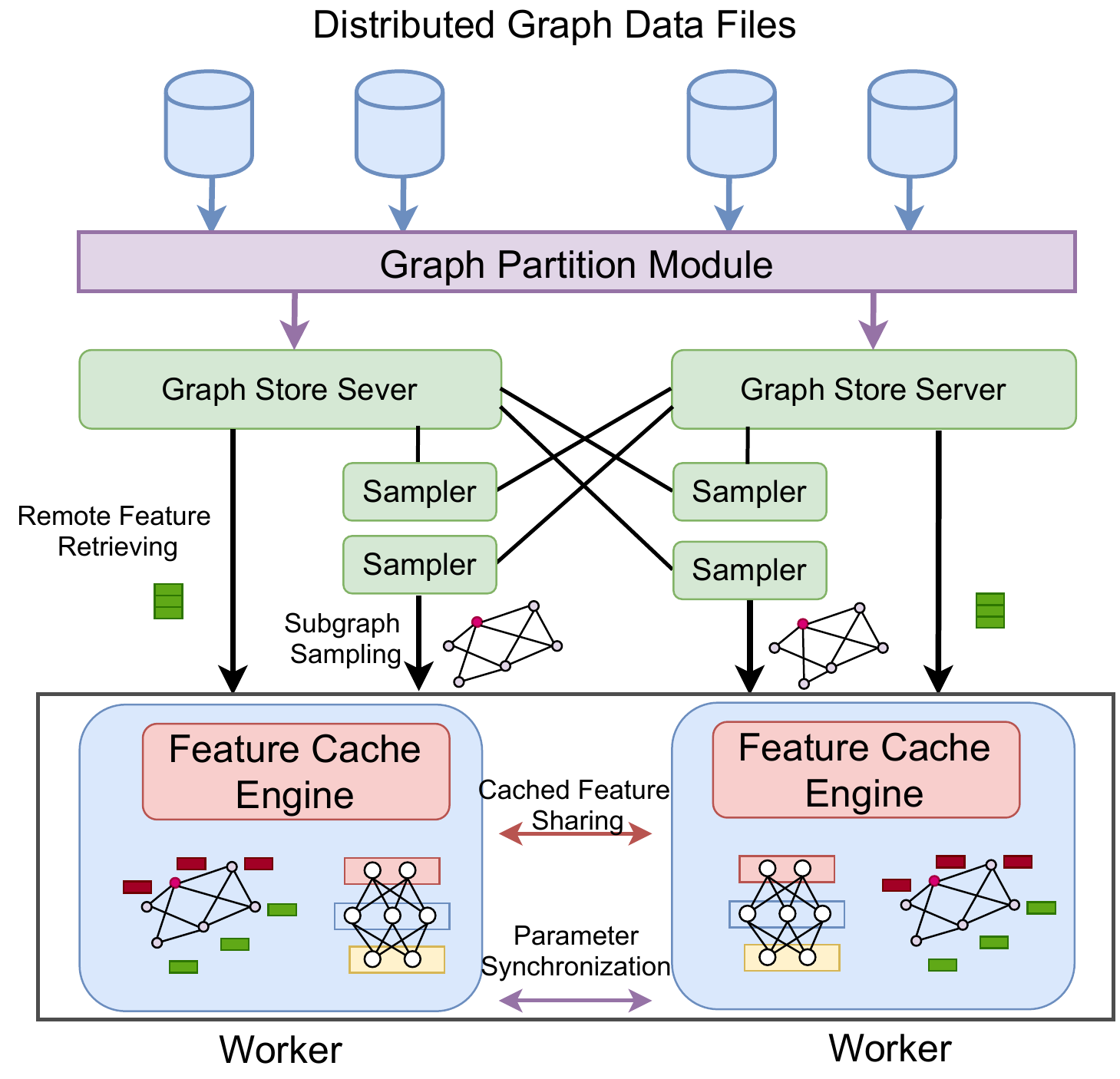}
\caption{The architecture of {\sysname}. }
\label{fig:work_flow}
\vspace{-5mm}
\end{small}
\end{figure}

%Raw graph data is stored in the distributed storage system (e.g., HDFS). 
The overall architecture of {\sysname} is shown in Figure~\ref{fig:work_flow}.
The \emph{graph partition module} loads the graph data stored in the distributed storage system ({\em e.g.,} HDFS), and shards the whole graph into several partitions.
Graph partitioning is a \emph{one-time cost}, and the results are saved in the distributed storage system used for different GNN training tasks.
Each partition can then be loaded into a graph store server's memory, ready for subgraph sampling. %\yrchen{Use another word, e.g., sampling request.}.
Samplers run on the CPUs of graph store severs.
%, responsible for sampling subgraphs. % for neighborhood information aggregation. 
They select %generate 
a number of training nodes and sample their multi-hop neighbors by iteratively sampling next-hop neighbors for several times.
If all the next-hop neighbors are stored in the current graph store server, samplers can get the list locally;
otherwise, they need to send network requests to other graph store servers. 
Each worker in {\sysname} runs on 1 GPU.
It receives sampled subgraphs from samplers and retrieves features of subgraph nodes from graph store severs, with a local \emph{feature cache engine} to improve the retrieving efficiency.
Furthermore, {\sysname} uses a fine-grained pipeline, allowing parallel and asynchronous execution of each stage.

\emph{To minimize the overhead of node feature retrieving},
{\sysname} proposes an algorithm-system co-design which can achieve high cache hit ratios with low dynamic cache overhead.
%On the system side, {\sysname} design a two-level cache engine containing GPU cache and CPU cache, to take full advantage of fast GPU memory and large CPU memory (\S \ref{sec:multi_level_cache}).
On the system side, {\sysname} designs a multi-GPU cache supporting dynamic caching policies.
To reduce the dynamic cache overhead, we use the FIFO policy (\S \ref{sec:multi_level_cache}) and carefully ensure the cache consistency among multiple GPU workers without locks (\S \ref{sec:cache_implementation}).
%Furthermore, it carefully decouples cache workflow to 
To achieve high cache hit ratios, on the algorithm side, samplers in {\sysname} select training nodes by \emph{proximity-aware ordering}.
This ordering algorithm uses BFS traversal to improve the temporal locality of subgraphs nodes in several consecutive mini-batches and carefully introduces randomness to preserve the state-of-the-art model accuracy (\S \ref{sec:proximity_aware_ordering}).
Under proximity-aware ordering, we show that simple FIFO can achieve highest cache hit ratios among popular complex polices, such as LRU and LFU.

\emph{To partition billion-node graphs for efficient sampling during GNN training}, in the graph partition module, {\sysname} first merges nodes into several blocks to reduce the graph size. 
We use multi-source BFS to preserve the multi-hop connectivity when merging blocks.
Preserving the connectivity of blocks, {\sysname} maximally assigns each block to the partition with the consideration of multi-hop connectivity, while optimizing the training workload balancing as well (\S \ref{sec:graph_partition_module}).

\emph{To allocate resources for contending pipeline stages}, % and improve resource utilization},
%{\sysname} isolates resources for contending pipeline stages, %based on the profiled results,
 by observing that some operations do not scale well with more resources under full contention, {\sysname} takes a resource isolation idea when assigning the resource for each pipeline stage. Specifically, {\sysname} formulates an optimization problem and assigns isolated resources accordingly to minimize the time of each pipeline stage under resource constraints.
(\S \ref{sec:resource_isolation})

\begin{figure}[!t]
\small
\begin{minipage}{\columnwidth}
	    \begin{subfigure}{0.49\linewidth}
			\centering
			\includegraphics[width=\linewidth]{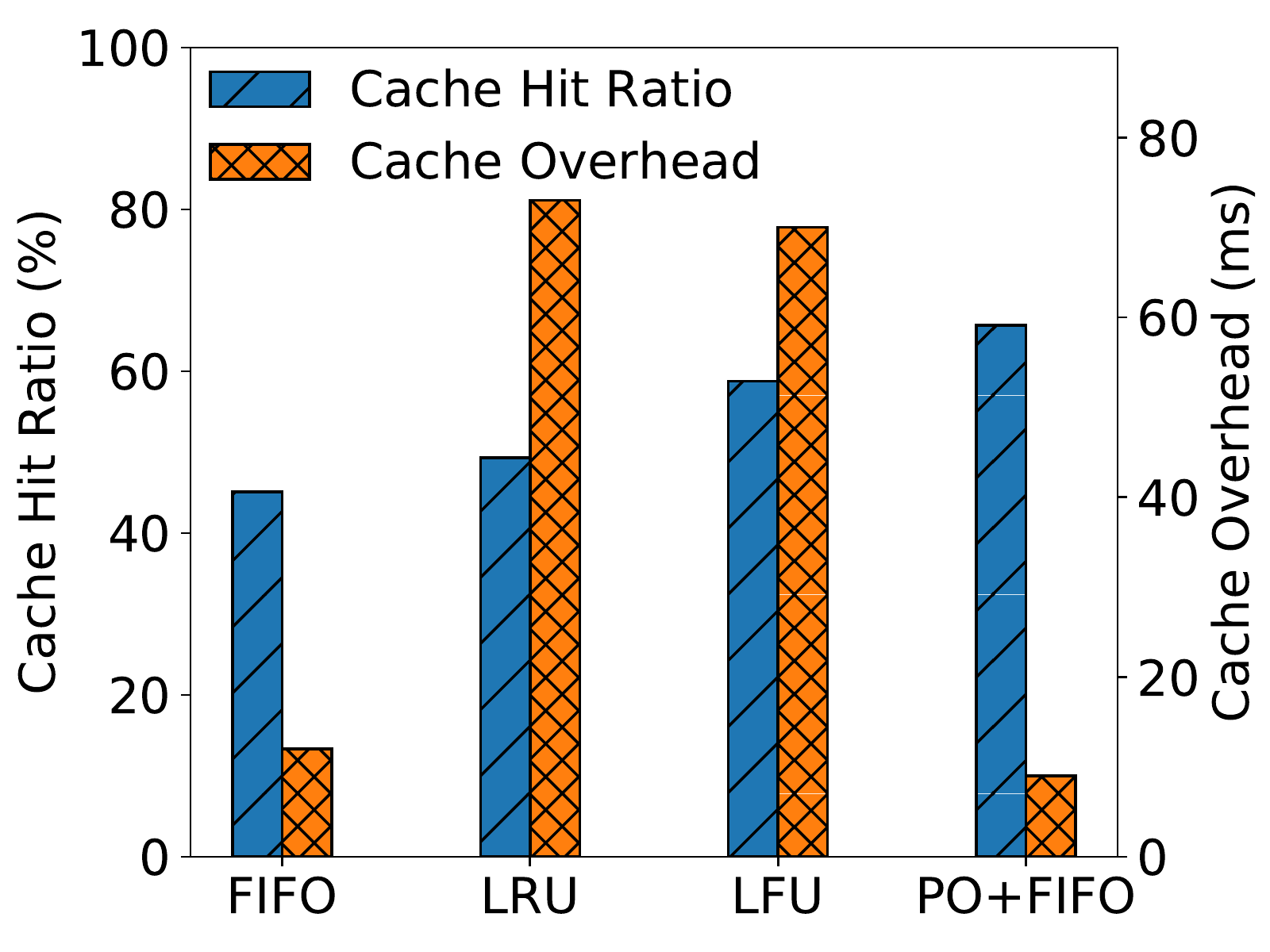}
			\caption{Trade-off between hit ratios and overhead (10\% cache size).}
			\label{fig:subfig_cache_trade_off}
		\end{subfigure}
		\hspace{1mm}
       	\begin{subfigure}{0.49\linewidth}
			\centering
			\includegraphics[width=\linewidth]{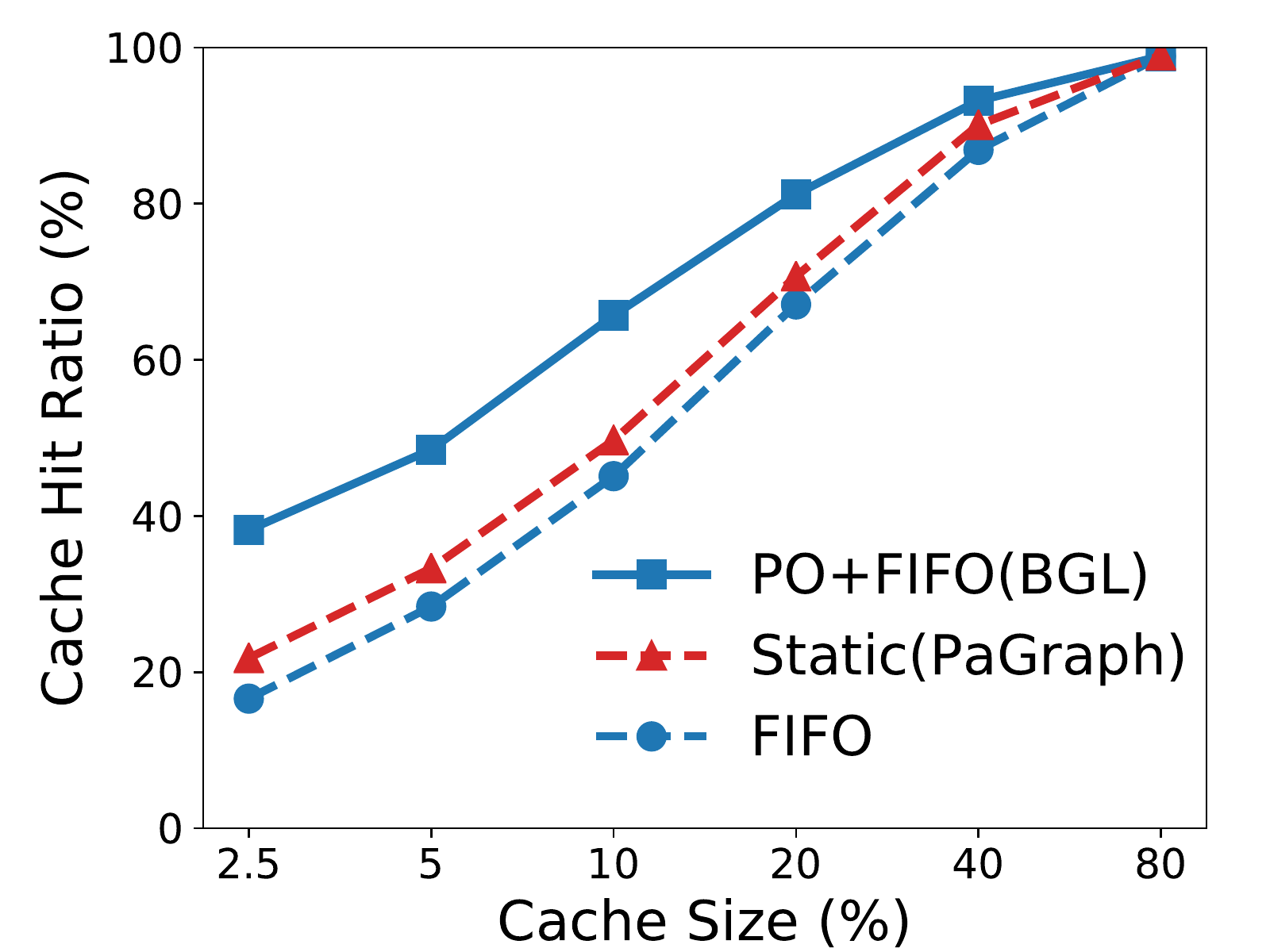}
			%\vspace{-3mm}
			\caption{Cache hit ratios with different cache sizes.}
			\label{fig:subfig_cache_hit_ratio}
		\end{subfigure}
		\vspace{-3mm}
		\caption{We test the cache hit ratios and overhead on Ogbn-papers with different cache sizes. PO is short for proximity-aware ordering, which is proposed in \S\ref{sec:proximity_aware_ordering}.}
		\label{fig:exp_cache_hit_ratio}
\end{minipage}

\vspace{-5mm}
\end{figure}

\subsection{Feature Cache Engine}
\label{sec:feature_cache_engine}

Feature retrieving contributes to a significant part of communication overhead in constructing a mini-batch. A feature cache can reduce the communication overhead upon a cache hit. However, the state-of-the-art design uses the static cache policy, which leads to low cache hit ratios for giant graphs.

\subsubsection{Dynamic Cache Policy}
\label{sec:multi_level_cache}

The first question is, which dynamic caching policy should we choose? PaGraph \cite{lin2020pagraph} indicates that dynamic policies have too high overhead. Based on our best-effort implementation,\footnote{We implement LFU and LRU with  $O(1)$ time complexity and use a contiguous 1D array as a hashmap to speed up key searching.}
we compare popular caching policies including LRU, LFU and FIFO in Figure~\ref{fig:subfig_cache_trade_off}. LRU~\cite{lru} and LFU~\cite{lfu} indeed have intolerable cache overhead.
FIFO's overhead (<20ms per batch) meets the throughput requirement for GNN training -- as mentioned in \S\ref{sec:mov}, an iteration of typical GNN model computation on GPU is around 20ms. In an asynchronous pipeline with cache as a part of data prefetching, FIFO cache will not become the bottleneck.

Note that in Figure~\ref{fig:subfig_cache_trade_off}, the overhead we report is the {\em amortized} time including cache lookup {\em and} cache update upon cache misses. Hence, a higher cache hit ratio, which means less frequent cache update for FIFO, can help to reduce the amortized overhead. We focus on amortized overhead because, again, cache is a stage in an asynchronous pipeline.

However, FIFO's cache hit ratio is unimpressive -- it is even lower than static policy's (Figure~\ref{fig:subfig_cache_hit_ratio}). %(Figure \ref{fig:exp_cache_hit_ratio}(b)). 
The reason is that FIFO does not leverage the distribution of node features. Regardless of how hot the node feature is, it is evicted as frequently as other colder node features.

\subsubsection{Proximity-Aware Ordering}
\label{sec:proximity_aware_ordering}
To address the above problem, we propose \emph{proximity-aware ordering} to improve the cache hit ratio with FIFO while maintaining low overhead. Figure~\ref{fig:subfig_cache_hit_ratio} %\ref{fig:exp_cache_hit_ratio}(b)
shows that FIFO combined with proximity-aware ordering can achieve the highest cache hit ratio among all candidate cache policies.

\begin{figure}[t]

\begin{minipage}{0.3\columnwidth}
			 \centering
    \includegraphics[width=0.78\linewidth]{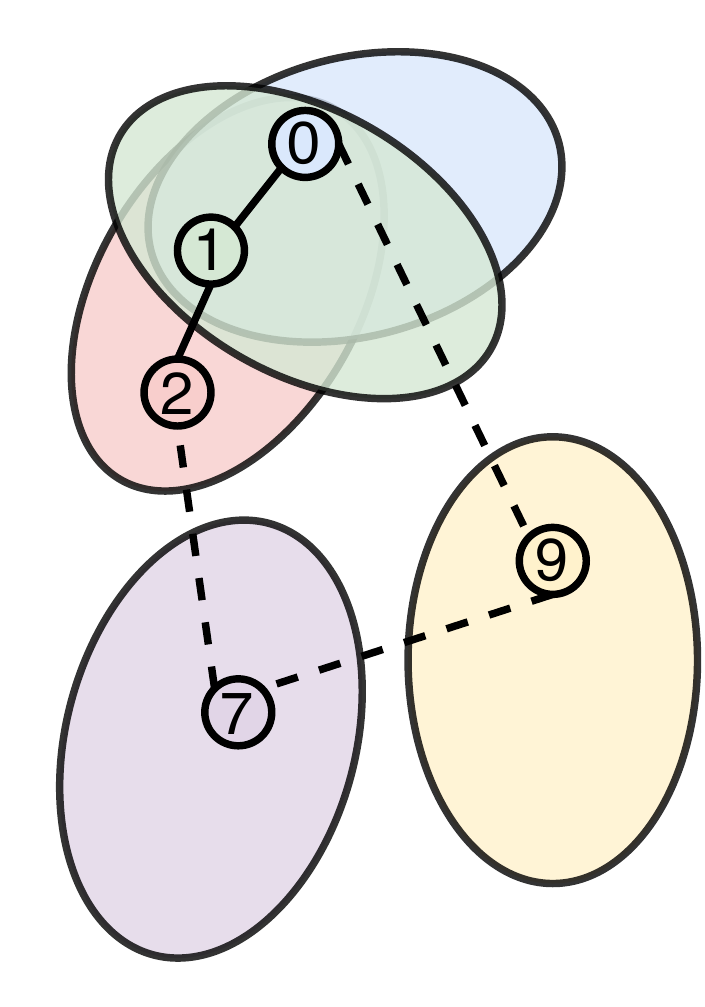}
    \vspace{-3mm}
    \caption{5 training nodes graph.}
    \label{fig:proximity_aware_ordering_intuition}
\end{minipage}
\hspace{3mm}
\begin{minipage}{0.65\columnwidth}
		 \centering
    \includegraphics[width=0.95\linewidth]{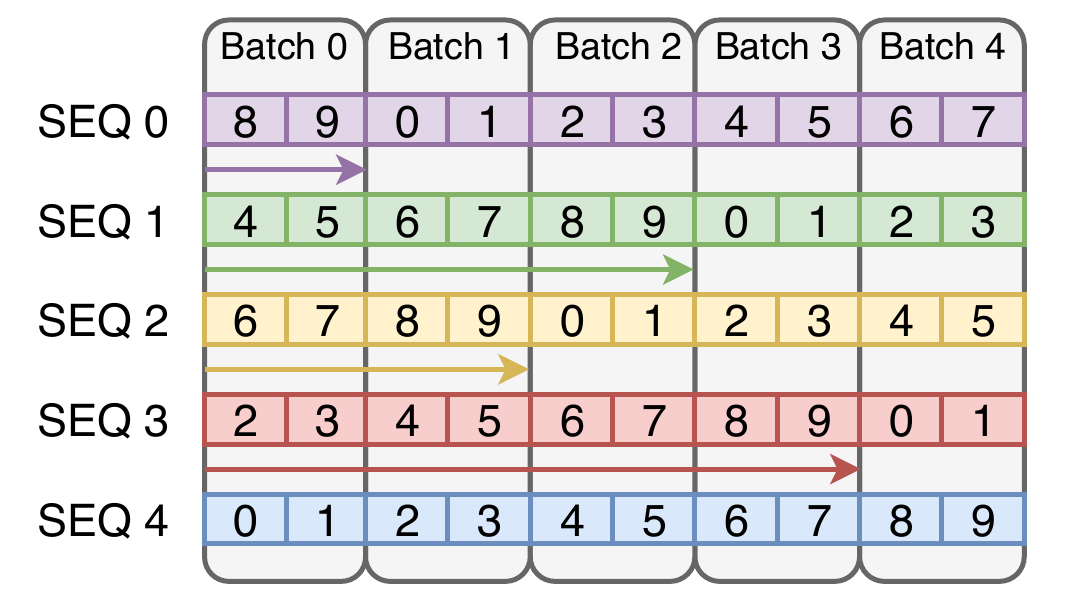}
    \vspace{-3mm}
    \caption{Two types of randomness introduced in proximity-aware ordering.}
    \label{fig:proximity_aware_ordering}
\end{minipage}
\vspace{-5mm}
\end{figure}

\new{We observe that each node may appear more than once among different training batches. This gives us an opportunity for data reuse by caching node features in nearby mini-batches (a.k.a., temporal locality). With random neighbor sampling, the chances of a node in nearby training batches are low. In order to increase cache hit ratio, we propose proximity-aware ordering by generating training nodes sequences in a BFS order. The ordering increases the probability that each node appears in consecutive batches and improves the cache hit ratio.
For example, there are five training nodes in Figure \ref{fig:proximity_aware_ordering_intuition}. Nodes 0, 1, 2 are direct neighbors (indicated by solid line) while nodes 7, 9 are remote nodes (indicated by dashed line).
Each colored circle represents the sampled subgraph of each node.
Assuming that the batch size is 1, the ordering \{0,1,2,7,9\} can maximize the temporal locality in consecutive batches, compared to random ordering (e.g., \{0,9,1,7,2\}). }

There is a trade-off between improving the temporal locality and ensuring model convergence.
Traversal-based ordering improves the temporal locality, but violates the i.i.d. requirement of SGD,  leading to different label distributions of batches and slow model convergence. 
On the other hand, random ordering, such as random shuffling, achieves state-of-the-art model accuracy by selecting random training nodes, with the cost of poor temporal locality.

Our proximity-aware ordering balances the above trade-off.
The key idea is that samplers still select training nodes based on the BFS traversal,  but we carefully introducing randomness.
As a result, if there is \emph{little difference} between the label distribution of each batch by proximity-aware ordering and that of all training nodes, the model convergence is kept.

We introduce the following randomness in proximity-aware ordering, as shown in Figure \ref{fig:proximity_aware_ordering}.
%Firstly, we use several different BFS sequences and generate them by selecting random BFS roots.
First, we use several different BFS sequences generated by selecting random BFS roots and select training nodes from different sequences in a round-robin manner to form a training batch.
Second, we randomly shift each BFS sequence.
We observe that the label distribution on the last several mini-batches are dramatically changed, because small connected components in giant graphs are more likely to be traversed at last and appended at the end of each BFS sequence.
Using random shifting not only randomizes this behavior but also preserves the order of consecutive nodes in BFS sequences.

\new{How many BFS sequences should we select?
We find, as long as the model convergence is guaranteed, we should use the minimum number of sequences to maximize the temporal locality. % \yrchen{The trade-off is not well presented.}
The theoretical theorem in \cite{meng2019convergence} shows that if there is \emph{little difference} between the output distribution of one algorithm $A$ and the uniform distribution, $A$ will not influence the convergence rate.
\cite{meng2019convergence} defines the difference $\epsilon$, named \emph{shuffling error}, as the total variation distance between the two distributions, and proves that, if $ \epsilon \leqslant \sqrt{bM}/{n}$, the convergence is not influenced, where $b$ is the batch size, $M$ is the number of workers and $n$ is the size of training data.}

Based on the above theorem, we determine the number of sequences as follows. %we can efficiently decide the number of sequences before GNN training.
We use the label distribution to calculate the shuffling error and estimate the probability of each label under the proximity-aware ordering as the frequency in per mini-batch.
%Due to the difficulty to estimate the node distribution using different BFS sequences, we .
{\sysname} firstly generates hundreds of BFS sequences and gradually increases the number of sequences from one until the shuffling error is smaller than the requirement of convergence (e.g., 5 sequences). Then, we generate five BFS sequences (SEQ0-SEQ4) at the beginning of each epoch and apply random shifting to each sequence (see Figure~\ref{fig:proximity_aware_ordering}). 

\begin{figure}[!t]
    \centering
    \includegraphics[width=0.85\linewidth]{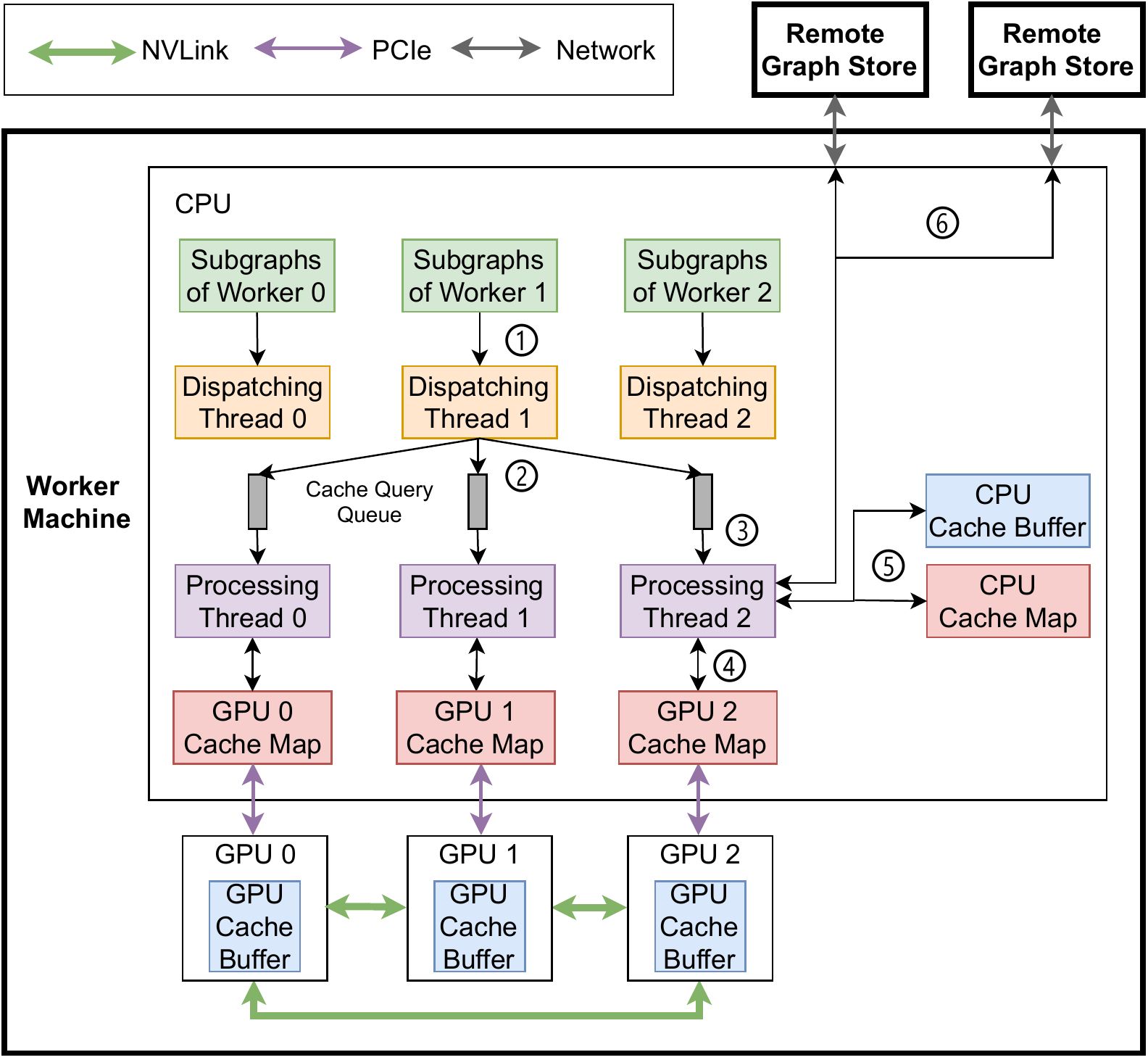}
\vspace{-3mm}
\caption{Structure and workflow of feature cache engine.}
\label{fig:multi_level_cache}
\vspace{-4mm} 
\end{figure}

\subsubsection{Maximizing Cache Size}
\label{sec:cache_implementation}
Orthogonal to the dynamic cache policy, another way to increase the cache hit ratio is to enlarge the cache size. In {\sysname}, we jointly use the memory of multiple GPUs (if the training job uses multiple GPUs) and CPU memory to build a two-level cache.
%{\sysname} propose a mutli-GPU cache supporting dynamic caching policies.
The detailed structure and cache workflow of our feature cache engine %on a worker machine 
are shown in Figure \ref{fig:multi_level_cache}.

\mypara{Multi-GPU Cache.}
As shown in Figure \ref{fig:multi_level_cache}, we create one {\em cache map} and one {\em cache buffer} for each GPU. 
Cache map is a HashMap with node IDs as keys and the pointers to buffer slots in cache buffer as values. Cache buffer consists of buffer slots, each of which stores a node feature.
Each GPU cache map manages its own GPU cache buffer.

To avoid wasting precious GPU memory, we ensure %that there are 
no duplicated entries among all GPU cache buffers by assigning different and disjoint node IDs to each GPU cache map (\texttt{mod} by the number of workers). A GPU can fetch node features from another GPU via point-to-point GPU memory copy. This works particularly well with GPUs that are interconnected by NVLinks, just like our NVIDIA V100-based environment.

Since CPU memory is much larger than GPU memory, {\sysname} also adds a CPU cache on top of the multi-GPU cache to further increase the cache size and reduce the communication traffic to graph store servers.
The CPU cache uses the same cache policy as the GPU cache, so we omit the details.

\mypara{Cache Workflow that Guarantees Consistency.}
As shown in Figure~\ref{fig:multi_level_cache}, the workflow of the cache engine goes as follows. 
After receiving sampled subgraphs  (\ding{192}), dispatching threads split the subgraph nodes by \texttt{mod} operation and send cache queries containing node IDs to \emph{cache query queues} (\ding{193}).
Each processing thread is assigned to each GPU cache buffer and processes all cache queries on this buffer (\ding{194}).
It first looks up the subgraph nodes in the GPU cache map, and then gathers cached features of those nodes %tensors 
from GPU cache buffers (\ding{195}).
In case of GPU cache misses, it looks up the CPU cache map for uncached nodes, gathers cached feature tensors from CPU cache buffer, and sends them to the GPU (\ding{196}).
The remainders are requested from graph store servers and sent to GPUs once they are received (\ding{197}). 
At last, the cache map and the cache buffer are updated according to our caching policies. 

There can be race conditions when the same cache buffer is read and written by different GPU workers simultaneously. A naive solution is to use locks. However, for GPUs, it means synchronization in CUDA APIs, leading to large overhead. Our solution is to queue all the operations towards a given GPU cache, including query and update. 
Only one processing thread polls the queue and then reads or writes the cache using CUDA. This reduces the overhead by 8x compared with using locks while avoids racing.

\begin{figure}[t]
\small
\centering
\includegraphics[width=0.8\linewidth]{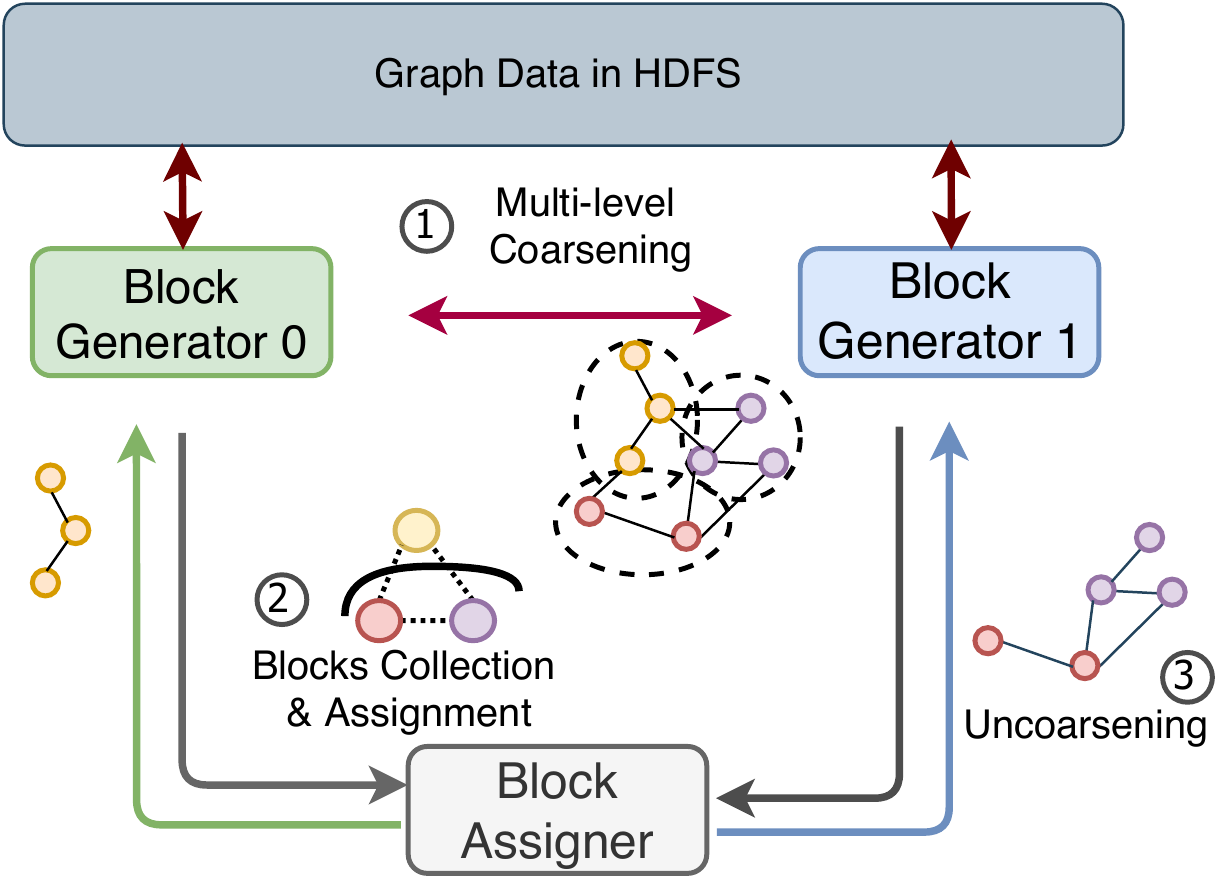}
\caption{The partition workflow of our partition algorithm.}
\label{fig:design-partition-framework}
\vspace{-3mm}
\end{figure}

\subsection{Graph Partition Module}
\label{sec:graph_partition_module}
 
Graph partition algorithms largely impact the performance of graph data sampling. 
As described in \S\ref{sec:mov}, a good partition algorithm should have the following properties: (i) Scalability to billion-node graphs, in terms of both memory and computation complexity,  (ii) Multi-hop connectivity: nodes in a multi-hop neighborhood can be accessed in a single or few requests. (iii) Load balance: the {\em training nodes} should be balanced across partitions. None of the existing algorithms has all the properties simultaneously.

Existing partition algorithm either fails to consider GNN's data accessing pattern or just unable to scale to web-scale graphs. 
To this end, we propose a novel graph partition algorithm. 
We make two key innovations: 
first, we adopt a multi-level coarsening strategy to greatly reduce the complexity of partitioning giant graphs while preserving multi-hop connectivity. 
\new{Second, we propose a novel assignment heuristic, which assigns nodes to partitions with multi-hop connectivity and training node load balancing to improve sampling speed. }

\subsubsection{Partition Workflow}

We outline the major steps of our partition algorithm in Figure~\ref{fig:design-partition-framework}.
A block is a connected subgraph that merges neighboring nodes for coarsening the graph, which is generated by block generators. 
We use multi-source BFS to merge nodes to preserving connectivity, which is the key difference from other algorithms, such as METIS.
The block assigner collects blocks from all block generators, and applies a heuristic (see \S\ref{sec:design_partition_assignment}) to assign each block to different partitions.The colors of nodes in the figure denote different blocks in the coarsened graph (step \ding{194} in Figure~\ref{fig:design-partition-framework}), or the nodes belonging to different blocks (step \ding{192} and \ding{193} in Figure~\ref{fig:design-partition-framework}).

\mypara{Coarsening:} 
We first perform multi-source BFS to divide the graph into many small blocks. 
Each block generator loads graph data from HDFS using a random partitioning algorithm, and randomly chooses a few source nodes within the partition for block merging.
\new{Each source node is assigned a unique Block ID and broadcasts the Block ID to its neighbors for generating blocks.}
A block is generated once the block size (\ie, the number of nodes with the same block ID) exceeds a threshold ({\em e.g.,} 100K), or there are no unvisited neighbors in BFS. When all nodes are visited, the block generating procedure stops.
Treating each block as one node, we obtain a coarsened graph from the original graph (see step \ding{193} in Fig~\ref{fig:design-partition-framework}). Each block generator maintains a \mbox{mapping} from the node ID to block ID, which is synchronized among all block generators for uncoarsening.

We further deploy a multi-level coarsening strategy to speedup the partition process.
 First, for small blocks connecting to large blocks~\footnote{Empirically, we set blocks with top 10\% sizes as large blocks.}, we merge them with their large block neighbors.
Second, other small blocks without large block neighbors are randomly merged.

The multi-level coarsening strategy scales well for billion-node graphs with numerous connected components~\cite{kwak2010twitter}. It significantly reduces the complexity of the following assignment heuristics as well as preserving the connectivity. 

\mypara{Assignment:} 
The block assigner collects the blocks of the multi-level coarsened graph from block generators for partitioning.
It applies a greedy assignment heuristics to form graph partitions, targeting both multi-hop locality and training node balancing. The block assigner then broadcasts the block partitions to the generators. We leave the details of the assignment heuristics in \S\ref{sec:design_partition_assignment}.

\mypara{Uncoarsening:} 
Upon receiving the block assignment results from the block assigner, the block generators start mapping back the blocks to the nodes in the original graph, \ie, uncoarsening. The partition results are then saved to the HDFS file (see step \ding{194} of Figure~\ref{fig:design-partition-framework}).

As a result, our partition algorithm has low time complexity and is friendly to gaint graphs. 
Let $\mathcal{E}_1$ be the set of edges in the coarsened block graph after BFS. $\mathcal{E}_2$ denotes the set of edges in the graph for assignment after multi-level block merging. With the multi-level coarsening, we reduce the time complexity of the assignment  %from $O(|\mathcal{E}|^j)$ ({\em e.g.,} PaGraph) 
to $O(|\mathcal{E}_2|^j)$, much lower than SOTA $O(|\mathcal{E}|^j)$~\cite{lin2020pagraph}, where $|\mathcal{E}_2| \ll |\mathcal{E}|$. The total partitioning complexity is $O(|\mathcal{E}| + |\mathcal{E}_1| + |\mathcal{E}_2|^j )$. 

\subsubsection{Assignment Heuristics}
\label{sec:design_partition_assignment}

\new{There is a trade-off when assigning nodes to partitions: a) on the one hand, nodes need to be assigned to the partition where most of their neighbors locate to reduce cross-partition communication when sampling; b) on the other hand, to balance the load among partitions, we also tend to assign nodes to the partition with larger capacity.}
To balance the trade-off, we propose the following heuristic for selecting the index of the partition where block $B$ is assigned:
\vspace{-2mm}
\begin{small}
\begin{align*} 
    \argmax_{i \in [k]} \left\{ \left( \sum_{j}  \left|P(i) \cap \Gamma^j(B)\right|\right) \cdot \left(1- \frac{|P(i)|}{C}\right) \cdot \left(1-\frac{|T(i)|}{C_T}\right)\right\},
\end{align*}
\end{small}
where $k$ is the number of partitions, each partition is referred by its index $P(i)$, and $B$ denotes the block to be assigned.

The heuristics contains three items.
The first item is the \emph{multi-hop block neighbor term}, $ \sum_{j} |P(i) \cap \Gamma^j(B)|$, which we tend to assign the current block to the partition with more multi-hop locality, where $\Gamma^j(B)$ refers to the set of $j$-hop neighbor blocks of $B$. 
The second item is the \emph{node penalty term}, $(1 - |P(i)|/{C})$, to balance the number of nodes in each partition, where $C = |\mathcal{V}| / k$  is the capacity constraint on each partition.
The third item is the \emph{training node penalty term}, $(1-|T(i)|/{C_T})$, to balance the training nodes across partitions, where $T(i)$ denotes the training nodes that have been assigned to $i$th partition, and $C_T = |\mathcal{T}| / k$ denotes training node capacity constraint on each partition.
We take multiplicative weights in this heuristics to enforce exact balance.

\begin{figure}[!t]
\begin{minipage}[t]{\columnwidth}
    \small
    \centering
    \includegraphics[width=\linewidth]{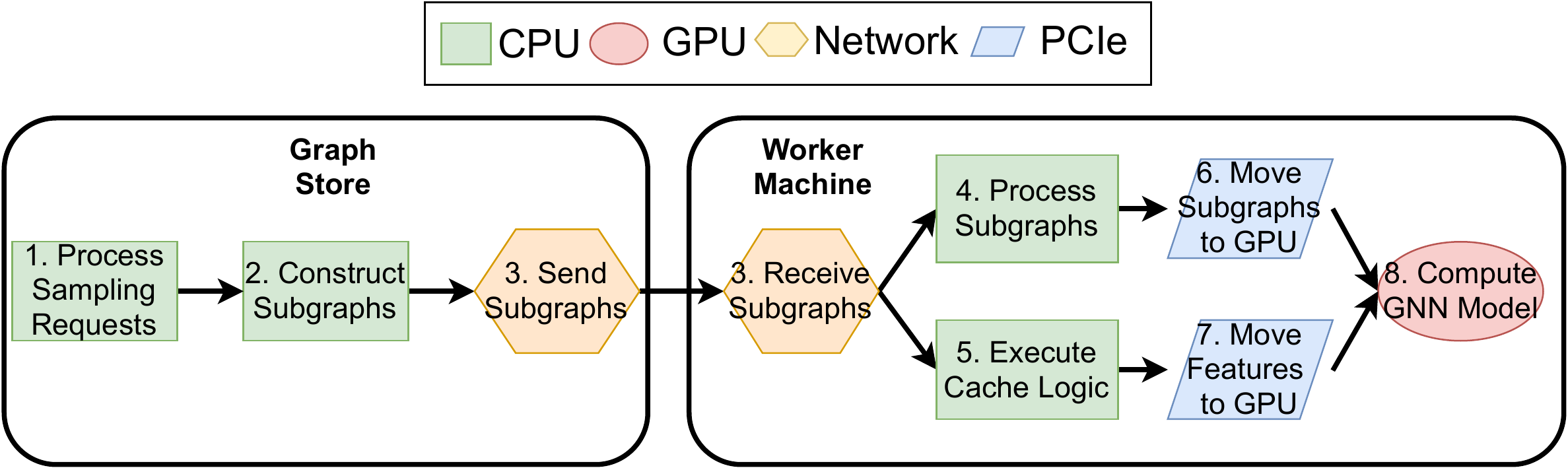}
    \caption{GNN training pipeline in {\sysname}.}
    \label{fig:dag_pipeline}
\end{minipage}
\vspace{-6mm}
\end{figure}

\subsection{Resource Isolation For Contending Stages}
\label{sec:resource_isolation}

To improve resource utilization and training speed, we divide GNN training into 8 asynchronous pipeline stages (see Figure~\ref{fig:dag_pipeline}) with careful consideration of data dependency and resource allocation. This is more complex than traditional DNN training. 
Some of the stages contend for resources on CPU, Network and PCIe bandwidth:
(i) Processing sampling requests and constructing subgraphs compete for CPUs on graph store servers.
(ii) Subgraph processing (e.g., converting graph format) %from COO to CSR)
and executing cache workflow compete for CPUs in the worker machine.
(iii) Moving subgraphs and copying features to GPUs compete for PCIe bandwidth.

However, we find that if all the processes freely compete for resources, the resource contention may lead to poor performance.
A key reason is that some stages do not scale well with more resources, while they may acquire more than they need for not becoming the bottleneck. 
For example, we observe that for the executing cache workflow stage (Stage 4 in Figure~\ref{fig:dag_pipeline}), when the number of CPU cores exceeds a threshold (e.g., 40), the performance converges or even degrades with more CPU cores (e.g., more than 64). This is because of the memory bandwidth limit and synchronization and scheduling overhead in the multi-threading library like OpenMP~\cite{bull1999measuring}.

To solve the above problem,
we propose a \emph{profiling-based resource allocation} to assign isolated resources to different pipeline stages.
We first profile the execution time of each stage, and then adjust resource allocation to balance the execution time of each stage.
We formulate the following optimization problem to compute the best resource allocation in a given GNN training task:
\begin{equation}
    \begin{aligned}
     \min{}\, & \max{} \left\{\frac{T_{1}}{c_{1}}, \frac{T_{2}}{c_{2}}, T_{net}, 
    \frac{T_3}{c_3}, \frac{D_{\RNum{1}}}{b_{\RNum{1}}},  f(c_4), \frac{D_{\RNum{2}}}{b_{\RNum{2}}}, T_{gpu}
    \right\} \\
     \textrm{s.t.} \, & \quad c_1 + c_2 \leqslant C_{gs} \\
                     & \quad c_3 + c_4 \leqslant C_{wm} \\
                     & \quad b_{\RNum{1}} + b_{\RNum{2}} \leqslant B_{pcie}
    \end{aligned} \notag
\end{equation}

The objective is to minimize the maximal completion time of all pipeline stages.
The constraints are resource capacity constraints for CPU on graph store servers, CPU on worker machines and PCIe bandwidth. %, respectively. % to minimize the DAG's execution time. 
The main decision variables are $c_i$, the number of CPUs required for the $i$th stage ($i \in \{1,2,3,4\}$) in Figure~\ref{fig:dag_pipeline}; and $b_i$, PCIe bandwidth of the $i$th stage ($i \in \{\RNum{1},\RNum{2}\}$).
All the other quantities are profiled by our system, including the time of the $i$th stage $T_i$, the data size of processed subgraphs $D_{\RNum{1}}$, and the data size of missed features $D_{\RNum{2}}$\footnote{We use the average data size when the cache is stable after several batches.}.
$C_{gs}$ and $C_{wm}$ denote the number of CPU cores on graph store servers and worker machines respectively, and $B_{pcie}$ is the PCIe bandwidth of the worker machines.

We assume linear acceleration of CPU execution, except on processing caching operation (Stage 4 in Figure~\ref{fig:dag_pipeline}). 
We introduce a fitting function $f(c_4) = a/c_4 + d$ to output the completion time of caching stage with a certain number of CPU cores $c_4$, where $a$ and $d$ are approximated by pre-running the task.

We use brute-force search to find the optimal resource allocation. To reduce the search space, we add integer assumptions on bandwidth variables $b_{\RNum{1}}$ and $b_{\RNum{2}}$. The time complexity is $O(C^2_{gs} + C^2_{wm} + B^2_{pcie})$ 
In average, our method spends less than 20ms on searching the best resource allocation strategy for GNN training pipeline.

\section{Implementation}
\label{sec:impl}

We implement {\sysname} with over 4,400 lines of C++ code and 3,300 lines of Python code. 
We reused the graph store module and GPU backend of the open-sourced Deep Graph Library (DGL v0.5~\cite{dmlc2020dgl, DBLP:journals/corr/abs-1909-01315}), and utilized the graph processing module of GMiner~\cite{DBLP:conf/eurosys/ChenLZYYC18} for partitioning. 
Our design can be applied to other GNN frameworks %as well 
with little change.   
We are collaborating with the DGL team to upstream our implementation.

\mypara{Feature Cache Engine.}
Cache workflow in feature cache engine contains several GPU operations, such as copying tensor from CPU memory to GPU memory and launching GPU kernels to copy tensor from/to other GPUs.
To make cache processing asynchronous,
we enqueue all cache GPU operations into a {\em dedicated} CUDA stream, 
and pre-allocate dedicated CPU memory as buffer and pin the memory.
Our cache engine uses CUDA Unified Virtual Addressing, and enables fast GPU P2P communication on each cache processing thread. 
The cache processing thread constructs a lightweight CUDA callback function and enqueues it into the CUDA stream. 
The callback function counts the number of finished cache queries and notifies workers when finishing all cache queries. 

To further expedite FIFO performance, BGL uses multiple OpenMP threads to execute FIFO concurrently.
We maintain a lightweight atomic variable $tail$ shared by all OpenMP threads to record the next column index of GPU cache buffer for insertion or eviction.
When inserting a new node, each OpenMP thread finds the next position by atomically increasing \texttt{tail} and the real position is \texttt{(tail+1)\%buffer\_size} (we treat each GPU cache buffer as a circular queue).
If this position has an old node, it evicts the old node from the GPU cache map.
Since node features are not changed during training, old node features are implicitly evicted by inserting new node features based on our cache workflow in \S 
 \ref{sec:cache_implementation}.

\mypara{Inter-Process Communication.}
We use separate processes for sampling, feature retrieving and GNN computation stages.
To minimize the IPC overhead, we use shared memory to avoid unnecessary memory copy among different processes.
Specifically, we use Linux Shared Memory and CUDA IPC to avoid unnecessary CPU and GPU memory copy, respectively.

\section{Evaluation}
\label{sec:eval}

\begin{table}[]
\centering
        \begin{minipage}[]{\columnwidth}
        \small
        \setlength{\tabcolsep}{8pt}
        \centering
        \caption{Datasets used in evaluation.}
        \vspace{-2mm}
        \label{tab:datasets}
        \begin{tabular}{l|l|l|l}
        \toprule
         & \textbf{Ogbn-}     & \textbf{Ogbn-}       & \multirow{2}{*}{\textbf{User-Item}}\\
         & \textbf{products}   & \textbf{papers}     & \\
        \midrule
        \textbf{Nodes} & 2.44M & 111M  & 1.2B\\
        \textbf{Edges}  & 123M & 1.61B  & 13.7B\\
        \textbf{Feature Dimension} & 100 & 128 & 96 \\
        \textbf{Classes} & 47 & 172 & 2 \\
        \textbf{Training Set} & 196K & 1.20M & 200M \\
        \textbf{Validation Set} & 393K &125K &  10M \\
        \textbf{Test Set} & 2.21M& 214K& 10M\\
        \textbf{Disk Storage} & 6 GB & 279 GB & 861 GB\\
        \textbf{Memory Storage} & 11 GB & 510 GB & 2 TB \\

        \bottomrule
        \end{tabular}
    \end{minipage}
\vspace{-5mm}
\end{table}

\subsection{Methodology}
\label{sec:method}
\noindent\textbf{Testbed.} 
We evaluate {\sysname} on a heterogeneous CPU/GPU cluster with 1 GPU server and 32 CPU servers.  %\cwu{complete}.
The GPU server has 8 Tesla V100-SMX2-32GB GPUs (connected by NVLinks), 96 vCPU cores and 356GB memory. Each CPU server has 96 vCPU cores and 480GB memory. All servers are inter-connected with 100Gbps bandwidth using Mellanox CX-5 NICs. The graph datasets are stored in HDFS.

\noindent\textbf{Datasets.} 
As show in Table Table~\ref{tab:datasets}, we train GNNs on three datasets with different sizes, including two public graph datasets: Ogbn-products~\cite{DBLP:journals/corr/abs-2005-00687} and Ogbn-papers~\cite{DBLP:journals/qss/WangSHWDK20}, as well as a proprietary web-scale graph dataset: User-Item.

\noindent\textbf{GNN Models.}
We evaluate {\sysname} with three representative GNN models: GCN (Graph Convolution Network)~\cite{DBLP:conf/iclr/KipfW17}, GAT (Graph Attention Network)~\cite{DBLP:conf/iclr/VelickovicCCRLB18} and GraphSAGE~\cite{DBLP:conf/nips/HamiltonYL17}.
We use the same model hyper-parameters as OGB leaderboards~\cite{ogb_leaderboard}, e.g., 3 layers and 128 hidden neurons per layer. 

\noindent\textbf{Mini-batch Sampling Algorithms.}
In our experiments, we use Neighbor Sampling~\cite{DBLP:conf/nips/HamiltonYL17}, 
 which is shown to achieve comparable model performance with full-batch graph training.\footnote{{\sysname} can also be applied to other vertex-centric GNN sampling algorithms, e.g., layer-wise sampling~\cite{DBLP:conf/iclr/ChenMX18} and random walk sampling~\cite{DBLP:conf/kdd/YingHCEHL18}. We omit the evaluation of other sampling algorithms since it is beyond our scope.} 
We set the mini-batch size to 1000, {\em i.e.,} each mini-batch contains 1000 sampled subgraphs and each subgraph contains one training nodes and its three-hop neighbors with fanout \{15,10,5\} in all experiments.

\begin{figure*}[!t]
    \centering
    \begin{minipage}[t]{0.3\linewidth}
    \centering
    \subcaptionbox{\small GraphSAGE (1.38x - 45.98x)} {\includegraphics[width=\linewidth]{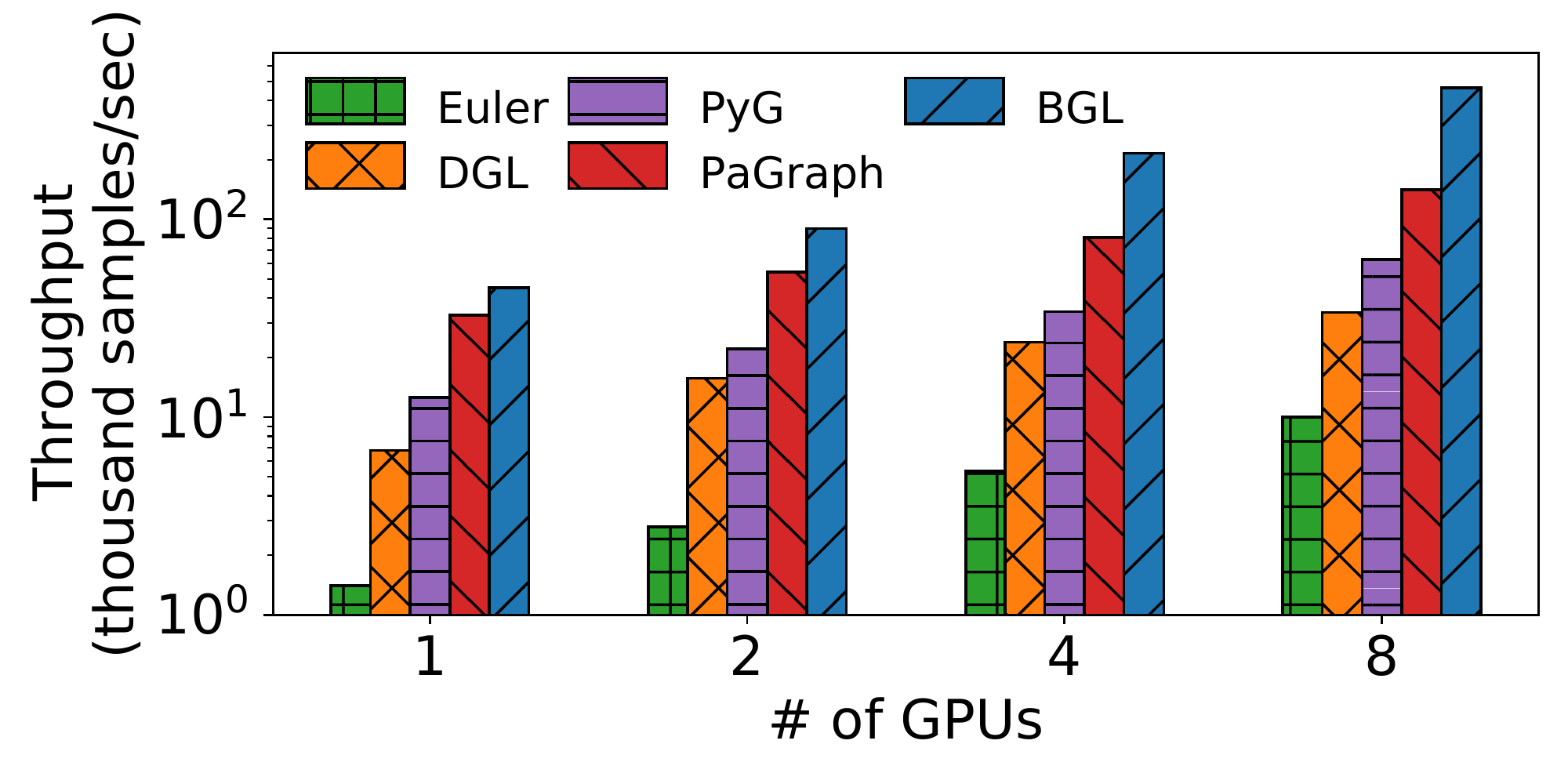}}
    \end{minipage}
    \begin{minipage}[t]{0.3\linewidth}
   	\centering
   	\subcaptionbox{\small GCN (1.49x - 41.33x)} {\includegraphics[width=\linewidth]{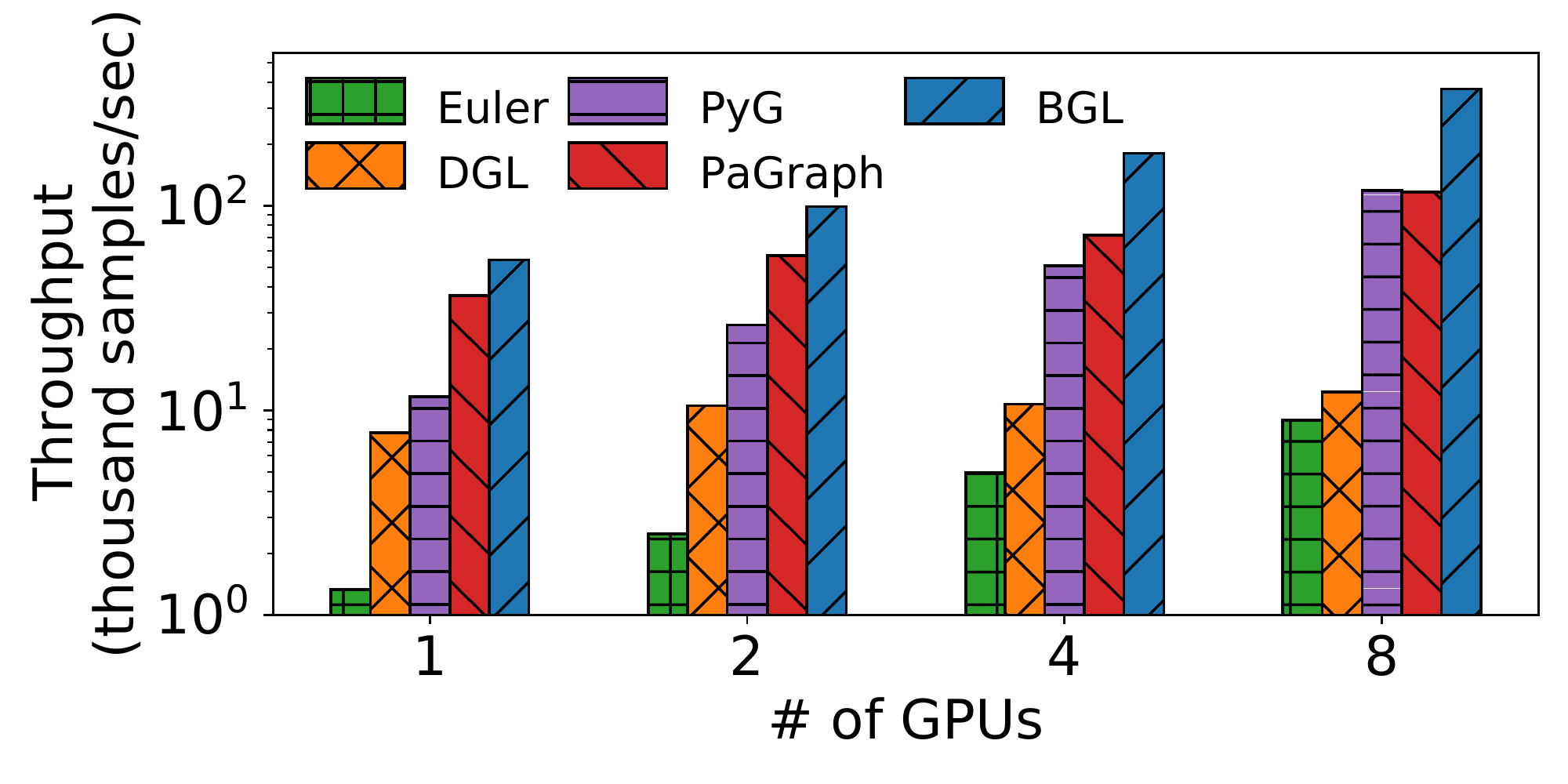}}
    \end{minipage}
    \begin{minipage}[t]{0.3\linewidth}
   	\centering
   	\subcaptionbox{\small GAT (1.14x - 40.60x)} {\includegraphics[width=\linewidth]{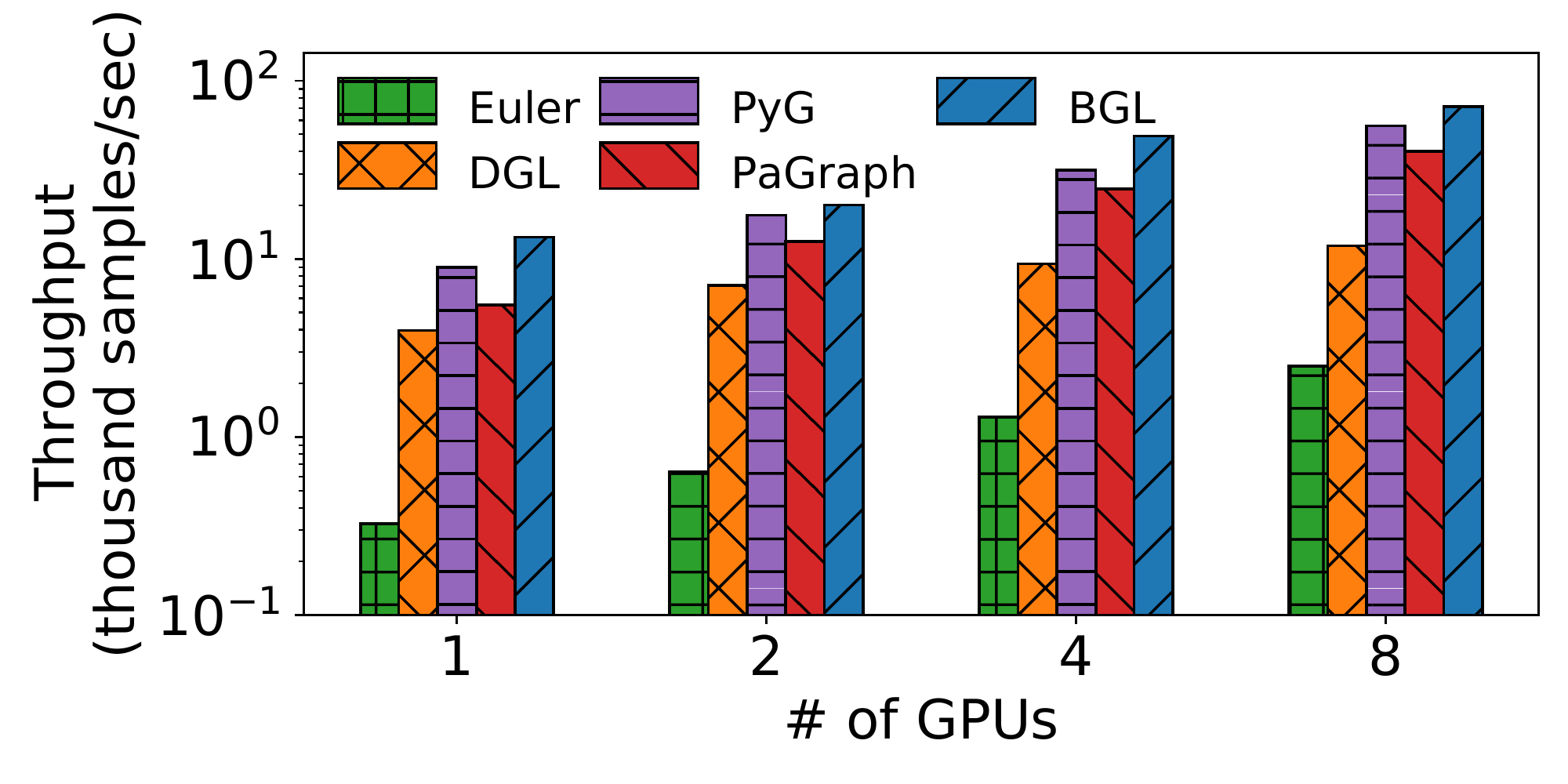}}
    \end{minipage}
    \vspace{-3mm}
    \caption{Training 3 GNN models over Ogbn-products (throughput in \textit{log} scale).}
        \label{fig:exp-overall-speed-product}
    \vspace{-2mm}
\end{figure*}

\begin{figure*}[!t]
    \centering
    \begin{minipage}[t]{0.3\linewidth}
    \centering
    \subcaptionbox{\small GraphSAGE (4.31x - 39.62x)} {\includegraphics[width=\linewidth]{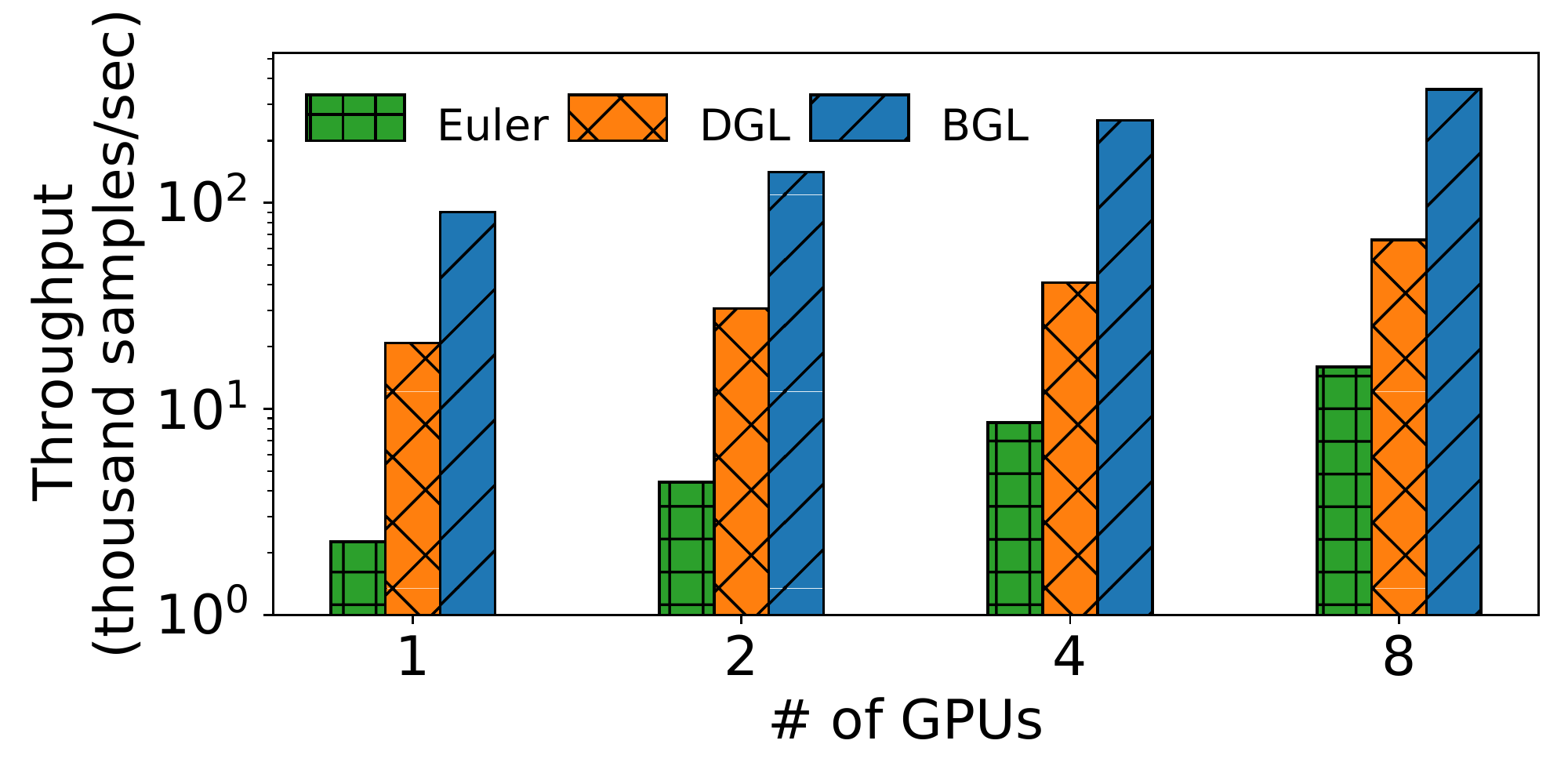}}
    \end{minipage}
    \begin{minipage}[t]{0.3\linewidth}
   	\centering
   	\subcaptionbox{\small GCN (4.62x - 32.11x)} {\includegraphics[width=\linewidth]{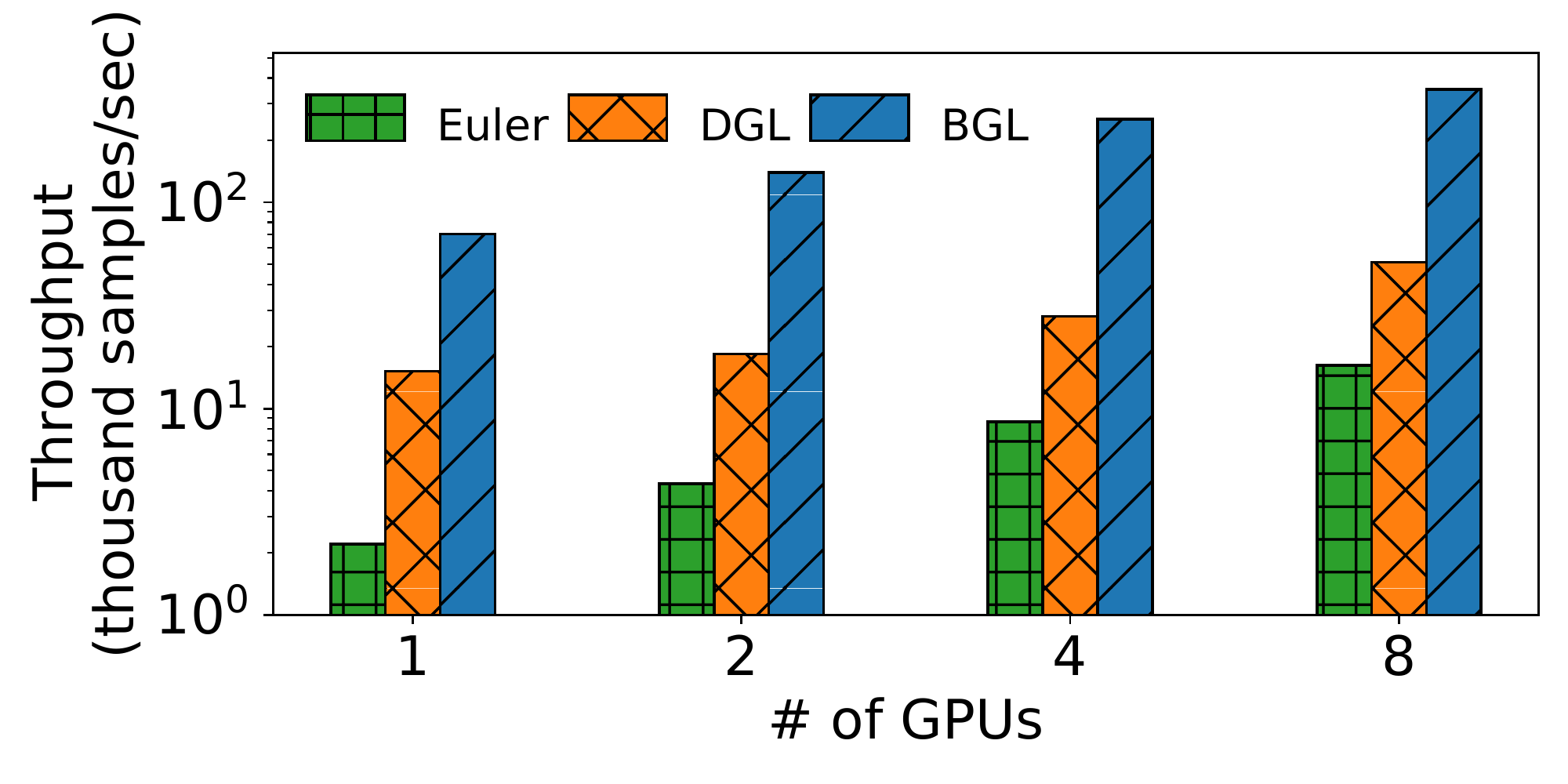}}
    \end{minipage}
    \begin{minipage}[t]{0.3\linewidth}
   	\centering
   	\subcaptionbox{\small GAT (1.26x - 68.97x)} {\includegraphics[width=\linewidth]{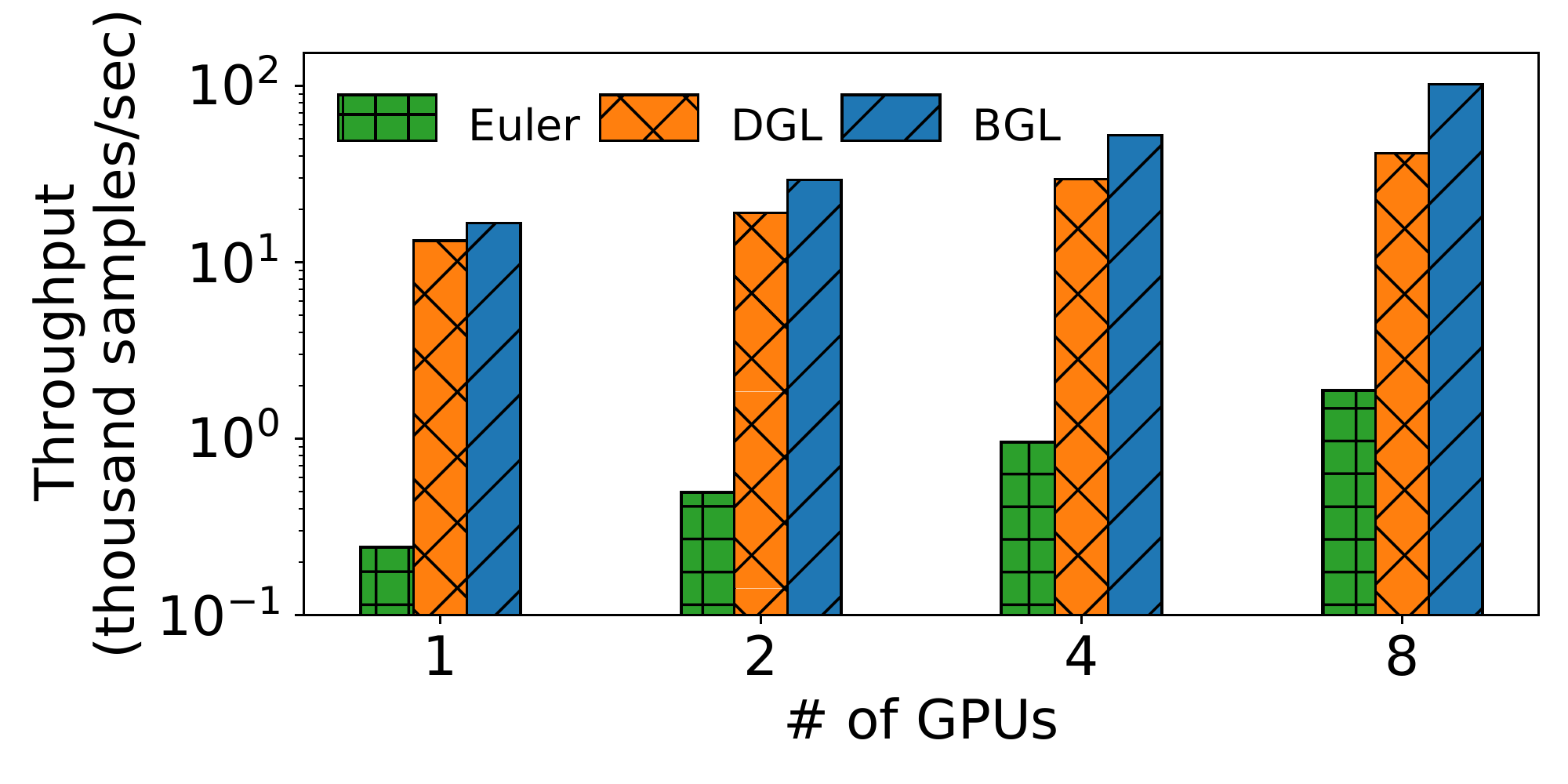}}
    \end{minipage}
    \vspace{-3mm}
    \caption{\begin{varwidth}[t]{\linewidth} Training 3 GNN models over Ogbn-papers (throughput in \textit{log} scale). We did not compare with PaGraph and \\ PyG because they fail to train Ogbn-papers (and User-Item) due to OOM or finish partitioning in 24 hours.
    \end{varwidth} }
    \label{fig:exp-overall-speed-paper}
    \vspace{-2mm}
\end{figure*}

\begin{figure*}[!t]
    \centering
    \begin{minipage}[t]{0.3\linewidth}
    \centering
    \subcaptionbox{\small GraphSAGE (3.02x - 14.16x)} {\includegraphics[width=\linewidth]{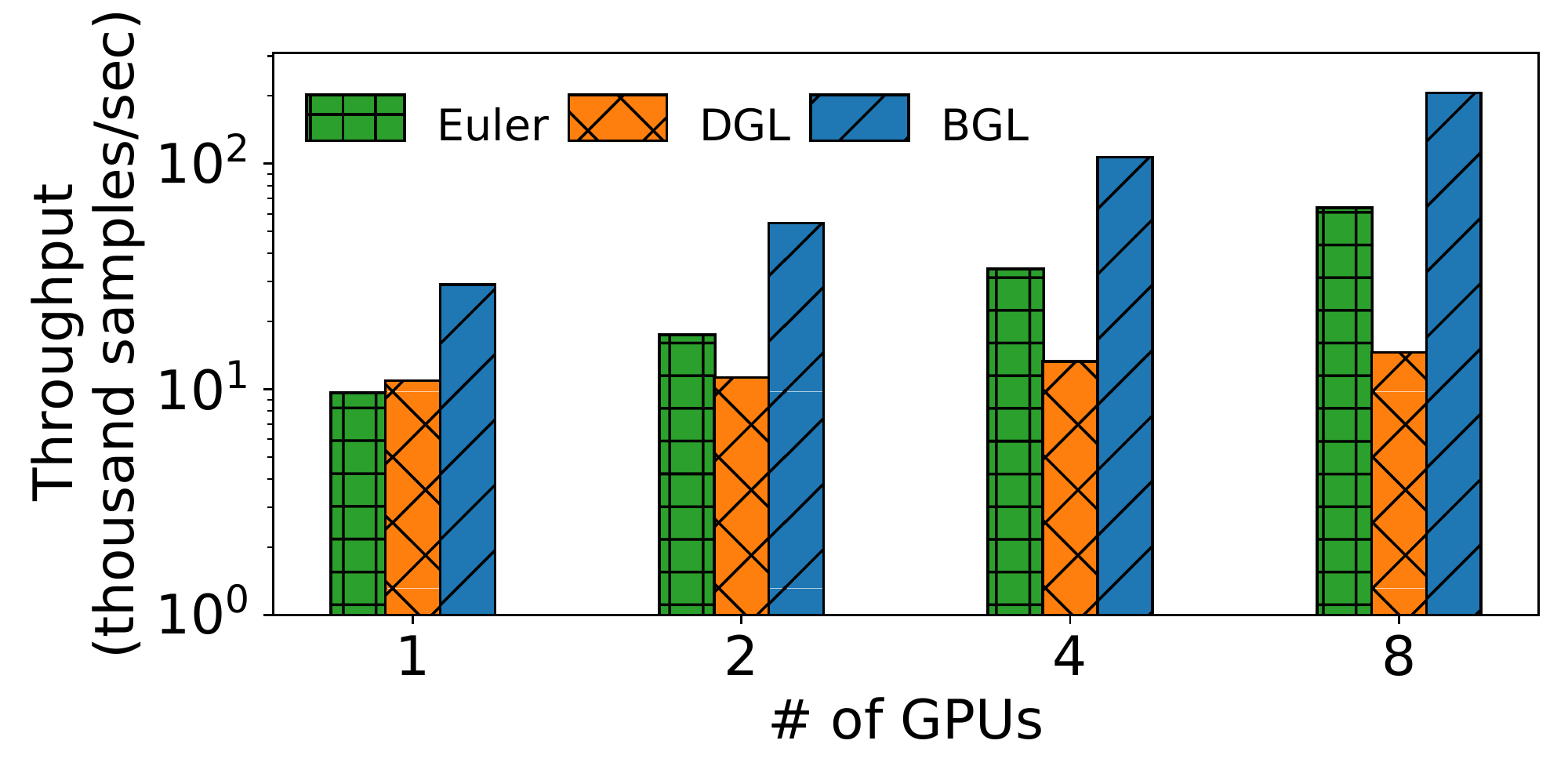}}
    \end{minipage}
    \begin{minipage}[t]{0.3\linewidth}
   	\centering
   	\subcaptionbox{\small GCN (2.97x - 12.08x)} {\includegraphics[width=\linewidth]{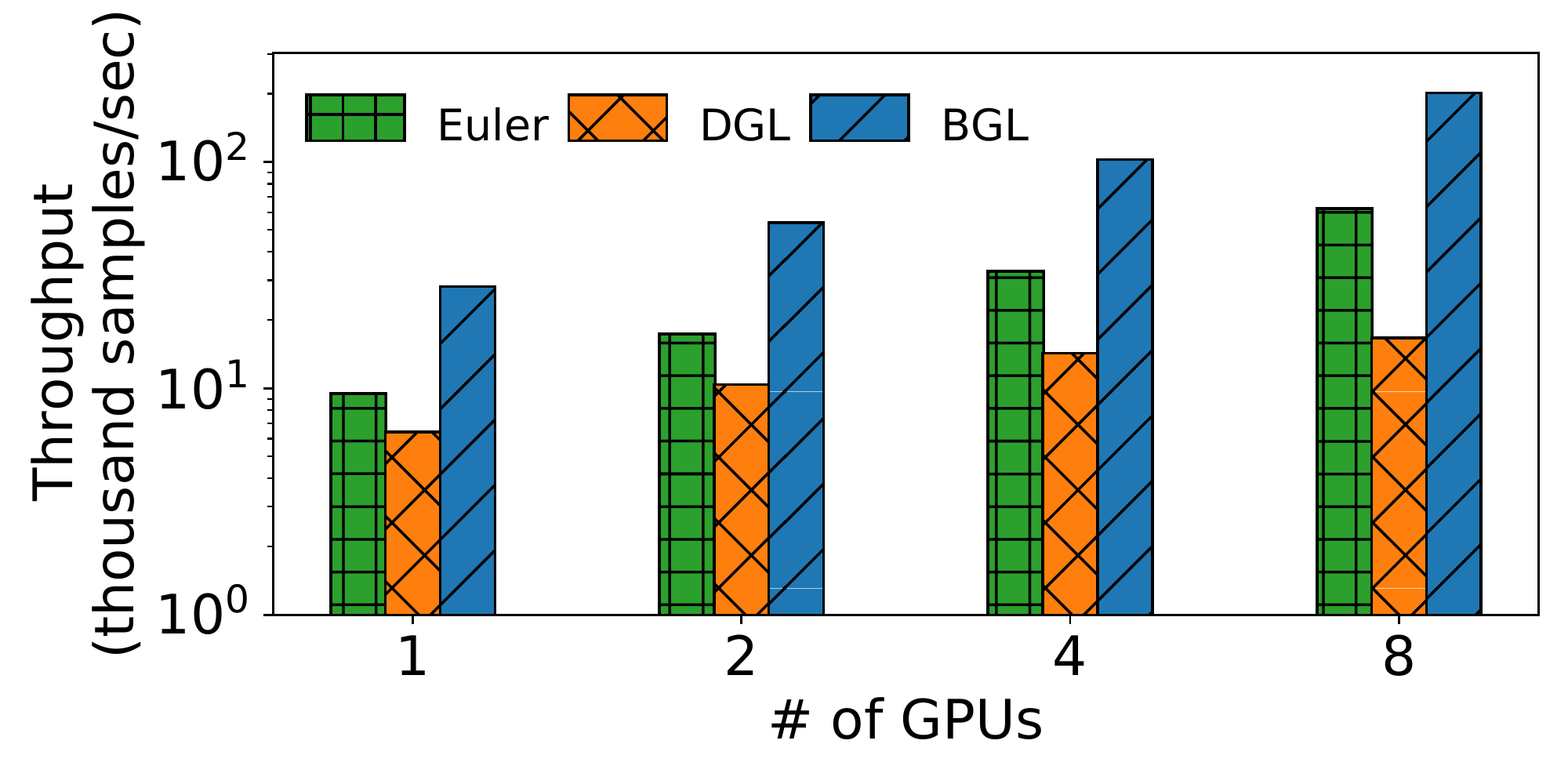}}
    \end{minipage}
    \begin{minipage}[t]{0.3\linewidth}
   	\centering
   	\subcaptionbox{\small GAT (2.86x - 29.73x)} {\includegraphics[width=\linewidth]{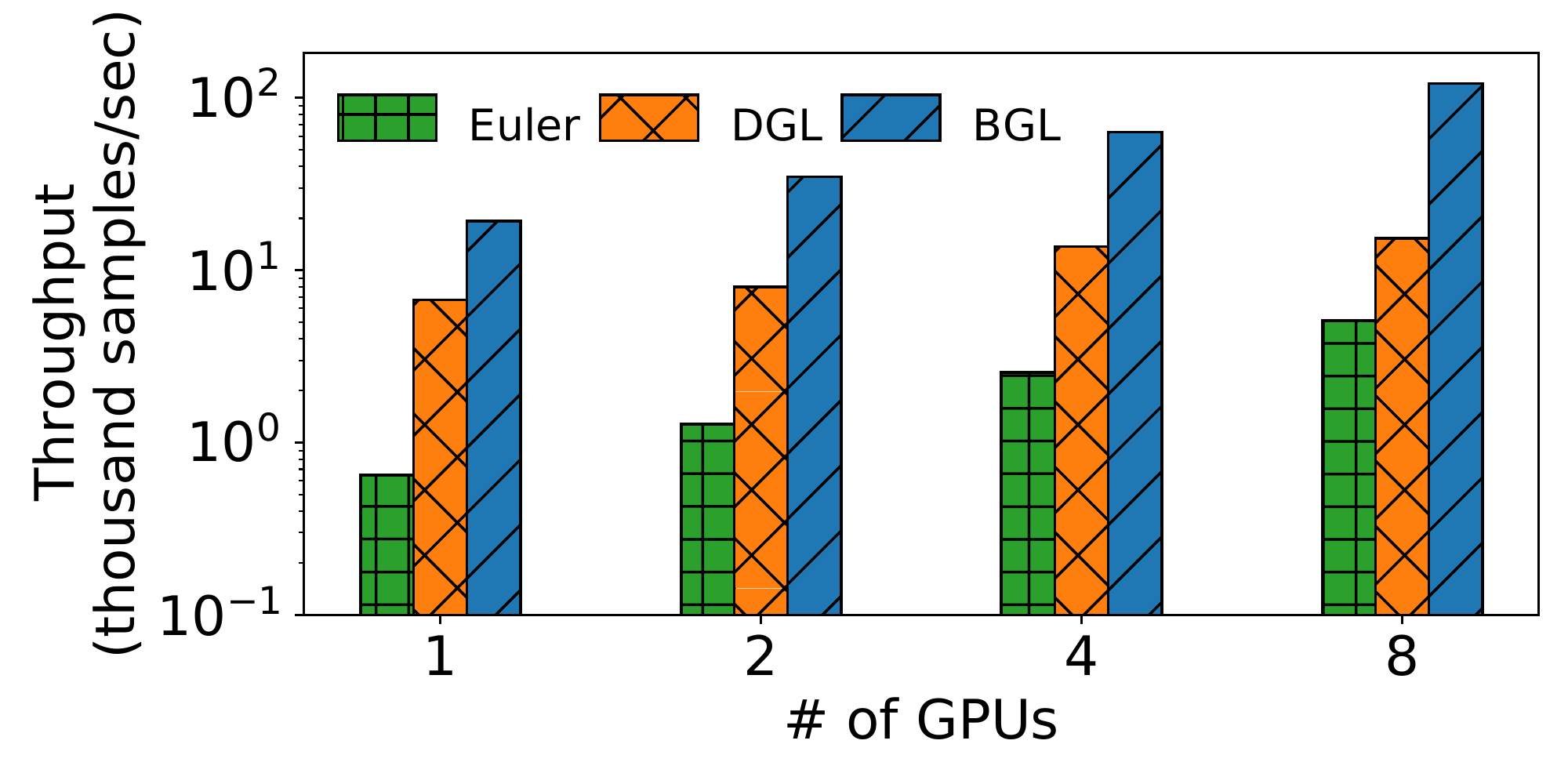}}
    \end{minipage}
    \vspace{-3mm}
    \caption{Training 3 GNN models over User-Item (throughput in \textit{log} scale).}
    \label{fig:exp-overall-speed-user-ad}
    \vspace{-5mm}
\end{figure*}

\noindent\textbf{Baselines.}
We use four open-sourced and widely-used GNN training frameworks as baselines for comparison \footnote{We omit $P^3$~\cite{gandhi2021p3} because it is not open-sourced.}.

$\bullet$ Euler~\cite{ali2019euler}: Euler (v1.0) 
is a distributed graph learning system built atop TensorFlow~\cite{DBLP:conf/osdi/AbadiBCCDDDGIIK16}.    We use TensorFlow's GPU backend for acceleration.
    
$\bullet$ DGL~\cite{dmlc2020dgl}: DGL is a deep learning library for graphs, compatible with multiple deep learning frameworks (e.g., Pytorch and TensorFlow). We use the DGL v0.5 release (DistDGL~\cite{zheng2020distdgl}).
    
$\bullet$  PyG~\cite{DBLP:journals/corr/abs-1903-02428}: PyG (v1.6.0) is an extension library for PyTorch, which consists of various methods for deep learning on graphs and a mini-batch loader for multi-GPU support in a single machine.
    
$\bullet$ PaGraph~\cite{lin2020pagraph}: PaGraph is a sampling-based GNN framework with a static cache strategy on GPU, which supports multi-GPU in a single server.

Specifically, PyG co-locates graph store servers and workers and allows graph sampling on the same machine only, making it unable to process large graph datasets (i.e., Ogbn-papers and User-Item) due to memory limit.
Also, PaGraph only supports generating partitions for each GPU in a single machine. Hence, we only compare {\sysname} with PyG and PaGraph on Ogbn-products dataset.
When training on User-Item dataset with DGL, we separate the graph store servers from the workers since our GPU servers do not have enough memory to load the graph partitions.
To evaluate the performance boundary, we use 32 and 8 CPU-based graph store servers for all frameworks on User-Item and Ogbn-papers, respectively.

\noindent\textbf{Graph Partitioning.}
DGL uses METIS partitioning for small graphs (\ie, Ogbn-products), and Random partitioning for large graphs that cannot be fitted into single machine (\ie, Ogbn-papers and User-Item).
Euler uses random partitioning for all graphs, and {\sysname} uses the proposed algorithm in \S \ref{sec:graph_partition_module}, where we set $j=2$, \ie, searching two-hop neighbors.

\subsection{Overall Performance}

Figure~\ref{fig:exp-overall-speed-product},~\ref{fig:exp-overall-speed-paper} and~\ref{fig:exp-overall-speed-user-ad} show the training speed of baselines and {\sysname} in \textit{log} scale when training the three GNN \mbox{models} on three graph datasets, with the number of workers ranging from 1 to 8, where each worker has one GPU. We use \linebreak \mbox{\textit{samples/sec}} as the metric to measure the training speed. A sample is a sampled subgraph of one training node.

\noindent\textbf{Different Frameworks.} {\sysname} achieves 1.14x - 69x speedups over four baselines in all settings. {\sysname} has 69x (the most) speedup over Euler. This is because Euler's random sharding in graph partition has very low data locality, resulting in frequent cross-partition communication in sampling. 
DGL does not cache features on GPU, introducing significant feature retrieving time. Thus, {\sysname} outperforms DGL by up to 30x.
PaGraph performs the best among baselines on Ogbn-product. It places graph structure data on each GPU with static caching on node features, leading to much faster data preprocessing.
Even in this case, {\sysname} still has up to 3.27x speedup, thanks to dynamic feature caching and  resource isolation for contending pipeline stages. 
{\sysname} outperforms all other systems, and the geometric mean of speedups over PaGraph, PyG, DGL and Euler is 2.14x, 3.02x, 7.04x and 20.68x, respectively.

\noindent\textbf{Different GNN models.} 
The training performance varies significantly across different GNN models. We see that {\sysname} achieves significantly higher performance improvement with GraphSAGE and GCN models, by up to 30x as compared to DGL and PyG.  
With the computation-intensive GAT model, however, the training speed of PyG and DGL is closer to that of {\sysname}. Hence the gain for {\sysname} ranges from 14\% to 8x.
It is because that the GAT model is computation-bound 
due to incorporating the attention mechanism into the propagation step, while its communication is less intensive than the other two GNN models; the higher ratio of computation over other stages results in smaller improvement space for {\sysname}. We see that Euler performs the worst in GAT, since it does not optimize the GPU kernels for irregular graph structures.

\noindent\textbf{Scalability.} 
{\sysname} also outperforms other frameworks in \linebreak terms of scalability. Without caching features on GPU, the throughput of baseline frameworks is bounded by PCIe bandwidth. For example, DGL has only 3x speedups when increasing the number of GPUs from 1 to 8. 
{\sysname} reduces the transmitted data through PCIe bandwidth 
with efficient GPU cache, resulting in linear scalability in throughput. Multi-GPU systems often suffer poor scalability due to synchronization overhead or resource contention. However, our design and implementation of multi-GPU memory sharing scales well with the increased number of GPUs.
With extra bandwidth brought by NVLink, accessing cache entries on other GPUs introduces negligible overhead. On the contrary, the increased cache capacity improved the cache hit ratio (see Figure~\ref{fig:subfig_cache_hit_ratio}) and overall feature retrieving time (see Figure~\ref{fig:exp_feature_retrieving_time}). 

\noindent\textbf{GPU Utilization.} We compare the GPU utilization achieved by {\sysname} and DGL with exactly the same GPU backend. We run GraphSAGE and GAT models on Ogbn-products dataset with 8 GPU.
{\sysname} achieves 99\% GPU utilization with the computation-intensive GAT model, while DGL's utilization is only 38\%. For GraphSAGE model with shallow neural layers, {\sysname} improves the GPU utilization from 10\%  to 65\%.

\begin{table}[!t]
    \centering
         \begin{minipage}[]{\columnwidth}
            \setlength{\tabcolsep}{3pt}
            \centering
            \caption{Graph sampling time (seconds) per epoch during training. {\sysname} reduces the graph sampling time by up to 25\%.  The number in brackets means the number of partitions in the corresponding setting.}
            \vspace{-2mm}
            \label{tab:exp_part_quality_compare}
            \begin{small}
            \begin{tabular}{c|c|c|c}
            \toprule
            & \textbf{Ogbn-products (2)} & \textbf{Ogbn-papers (4)} & \textbf{User-Item (4)}     \\
            \midrule
            Random & 66 & 252 & 643 \\
            GMiner  & 58 & 209  & 931 \\
            {\sysname}  & 50 & 187 & 519 \\
            \bottomrule
            \end{tabular}
            \end{small}
        \end{minipage}
\vspace{-4mm}
\end{table}

\subsection{Individual Components}
\subsubsection{Graph Partition}
\label{sec:eval_ablation_partition}

We compare the graph partition algorithm in {\sysname} with Random and GMiner partitioning, since only these two partition algorithms can scale to Ogbn-papers and User-Item.
We evaluate the sampling time under different partition algorithms and the one-time partition time (counted from loading the graph data to saving the partition results to files).

Table~\ref{tab:exp_part_quality_compare} shows the graph sampling time (per epoch) under different partition algorithms. {\sysname} achieves the best performance across different graph datasets, reducing the sampling time by at least 20\% %up to \jun{up to or at least..}25\%
over random partition algorithm. {\sysname} significantly reduces the cross-partition communication in distributed neighbor sampling by including the multi-hop locality when partitioning. Compared to GMiner, which also preserves connectivity,  {\sysname} manages to drop the sampling time by 14\% and 10\% for Ogbn-products and Ogbn-papers, respectively, thanks to training node balancing and multi-hop connectivity of partitioning. %\jun{here the difference is only node balancing ? what about two-hop and Gminer only has 1-hop} 
It is worth noting that, in some cases, GMiner even results in longer graph sampling time than random partitioning algorithm (e.g., in User-Item), showing the disadvantage of imbalanced sampling workload. %\jun{last sentences implies random partition has node load balancing, can you confirm ?} %\yrchen{Yes, random partition has node load balancing.}

{\sysname} introduces multi-level coarsening to mitigate the extra complexity brought by computing two-hop locality. Table~\ref{tab:exp_part_time_compare} shows {\sysname}'s partition algorithm runs as fast as the well-optimized original GMiner.

\begin{table}[!t]
    \centering
    \begin{minipage}[]{\columnwidth}
        \setlength{\tabcolsep}{3pt}
        \centering
        \caption{One-time partition time cost (seconds) before training. This partition cost can be amortized by lots of training tasks, each containing hundreds of epochs to converge. The number in brackets has the same meaning as Table \ref{tab:exp_part_quality_compare}.} % The number in brackets means the number of partitions in the corresponding setting.}
        \vspace{-2mm}
        \label{tab:exp_part_time_compare}
        \begin{small}
        \begin{tabular}{c|c|c|c}
        \toprule
        & \textbf{Ogbn-products (2)} & \textbf{Ogbn-papers (4)} & \textbf{User-Item (4)}     \\
        \midrule
        %Random & 21 & 208 & 691 \\
        GMiner  & 38.2 & 1636  & 10679 \\
        {\sysname}  & 23.5 & 1607 & 8468 \\
        \bottomrule
        \end{tabular}
        \end{small}
    
    \end{minipage}
\vspace{-5mm}
\end{table}

\subsubsection{Feature Cache Engine}
\label{sec:eval_feature_cache_engine}
In \S\ref{sec:feature_cache_engine}, we have shown the cache hit ratio with different cache policies and cache sizes. Here, we present the amortized feature retrieving time with the feature cache engine. %in real training workloads.

We compare the feature retrieving time of one mini-batch using different GPUs on Ogbn-papers.
PaGraph cannot scale to such large graphs. Therefore we implement its static caching policy in {\sysname}, which caches the features of high-degree nodes.
Euler and DGL do not have cache, so the feature retrieving time is the elapsed time of transmitting features from graph store servers to GPU memory.
As shown in Figure~\ref{fig:exp_feature_retrieving_time} (in log scale), due to high cache hit ratios and low cache overhead, the feature retrieving time of {\sysname} is the shortest among all systems.
Compared to other systems on 1 GPU worker, {\sysname} reduces the feature retrieving time 98\%, 88\% and 57\% for Euler, DGL and PaGraph respectively. %\yrchen{Show the exact improvement number over PaGraph.}

\subsubsection{Resource Isolation}
\label{sec:eval_adaptive_resource_allocation}

To evaluate the effectiveness of our resource isolation mechanism, we compare {\sysname} with Euler, DGL and {\sysname} without resource isolation when training GraphSAGE with 4 GPUs on datasets Ogbn-products and Ogbn-papers.  
``BGL w/o isolation'' is a naive resource allocation method, which shares all resources among pipeline stages. It increases resource utilization but incurs larger contention and parallel overhead.

\begin{figure}[!t]
\begin{minipage}{0.46\columnwidth}
			\centering
			\includegraphics[width=\linewidth]{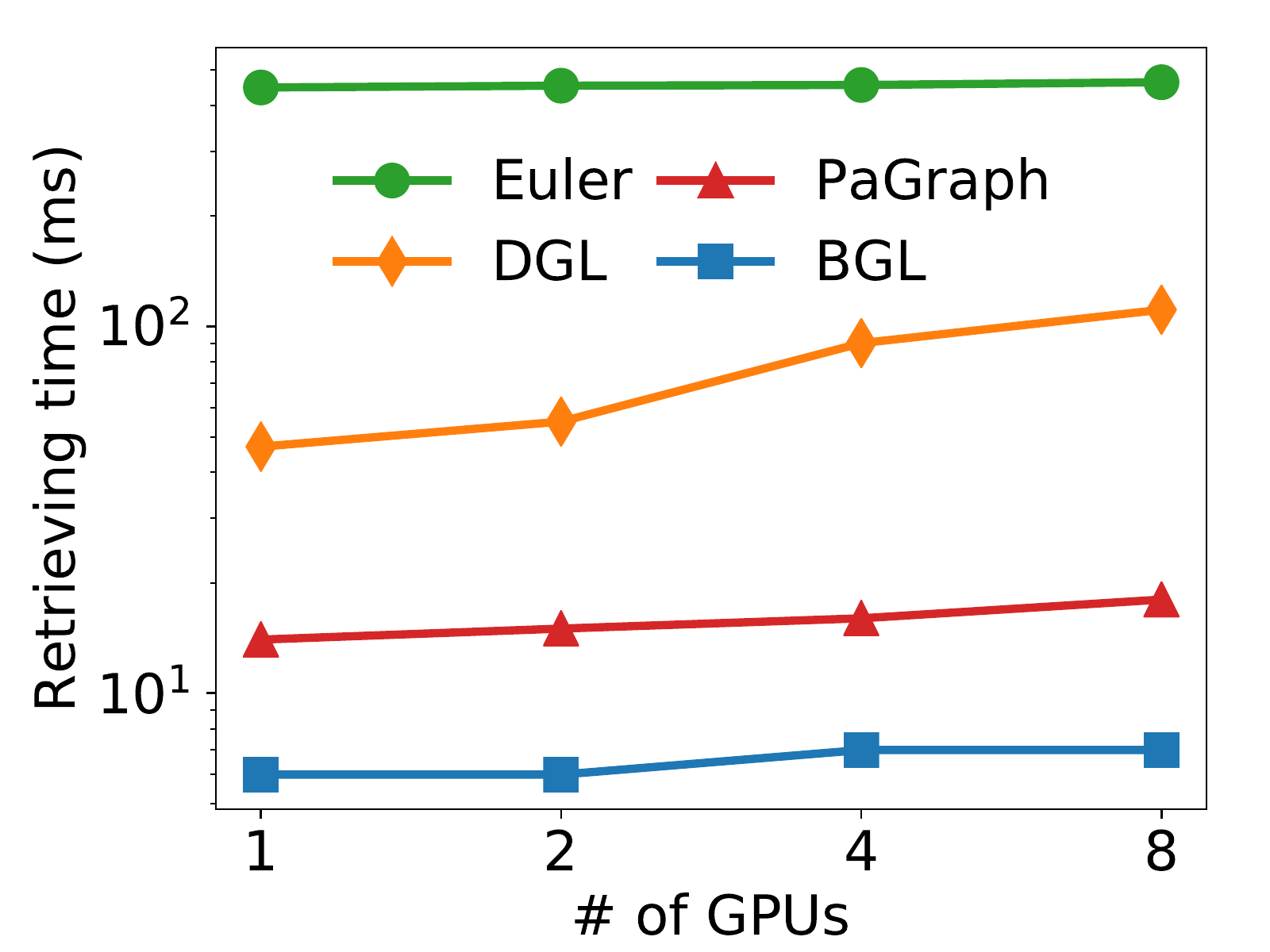}
		\vspace{-5mm}
		\caption{Feature retrieving time of one mini-batch on Ogbn-papers.}
		\label{fig:exp_feature_retrieving_time}
\end{minipage}
\hspace{3mm}
\begin{minipage}{0.46\columnwidth}
			\centering
			\includegraphics[width=\linewidth]{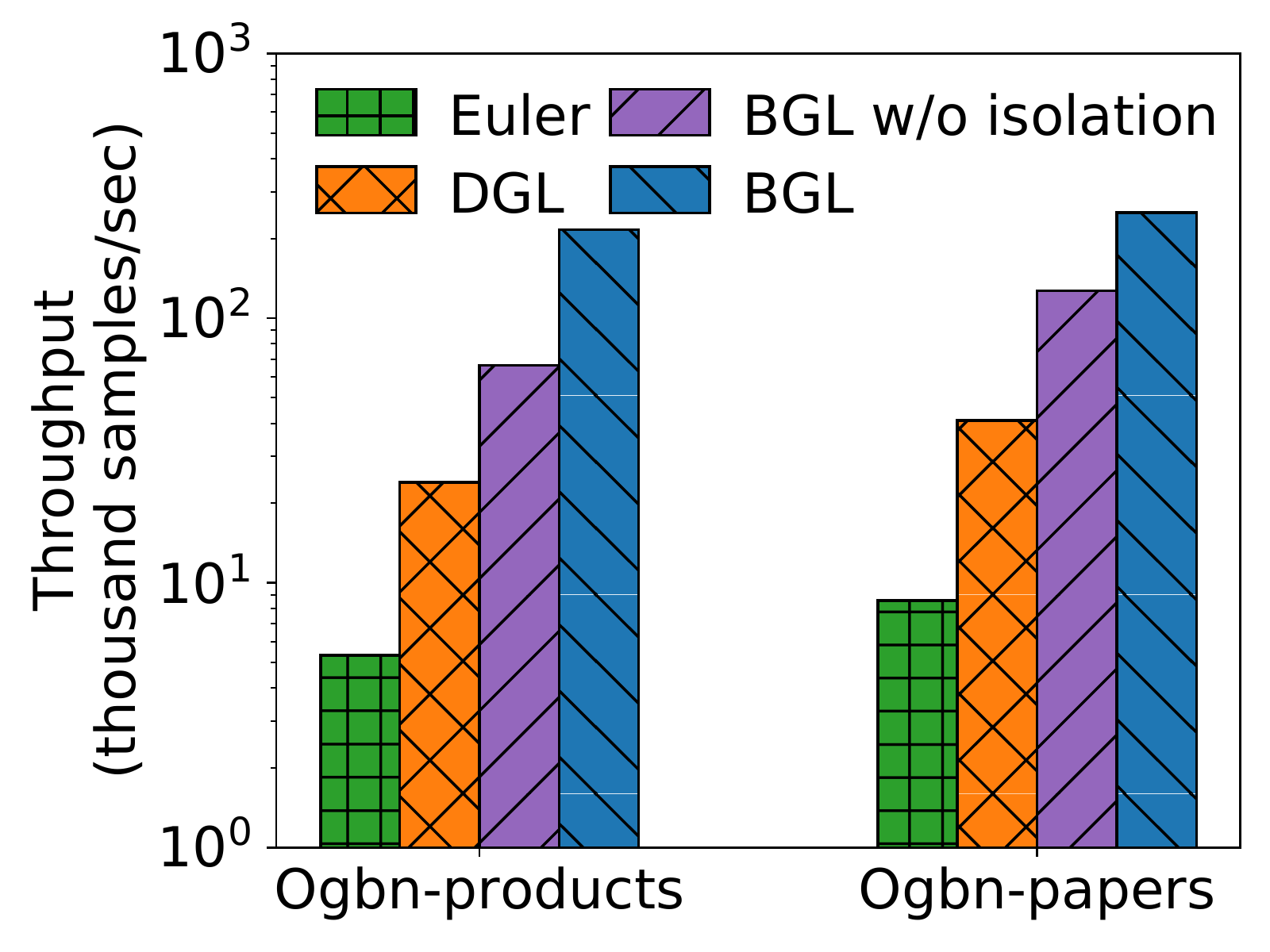}
		\vspace{-5mm}
		\caption{Training throughput on GraphSAGE using 4 GPUs. %"BGL w/o isolation" is {\sysname} without resource isolation.
		}
		\label{fig:exp-pipeline-iter-time}
\end{minipage}
\vspace{-2mm}
\end{figure}

As shown in Figure~\ref{fig:exp-pipeline-iter-time} (in log scale), {\sysname} achieves the highest throughput across all settings. Both {\sysname} and ``BGL w/o isolation'' outperform the baselines, namely, Euler and DGL. As compared to the naive resource allocation strategy without isolation, {\sysname} speedups the throughput by up to 2.7x. It mitigates the resource contention among different pipeline stages, and incurs lower parallel overhead of OpenMP.

\subsection{Model Accuracy}
\label{sec:eval_convergence}
\begin{figure}[t]
\begin{minipage}{\columnwidth}
       	\begin{subfigure}{0.45\linewidth}
			\centering
			\includegraphics[width=\linewidth]{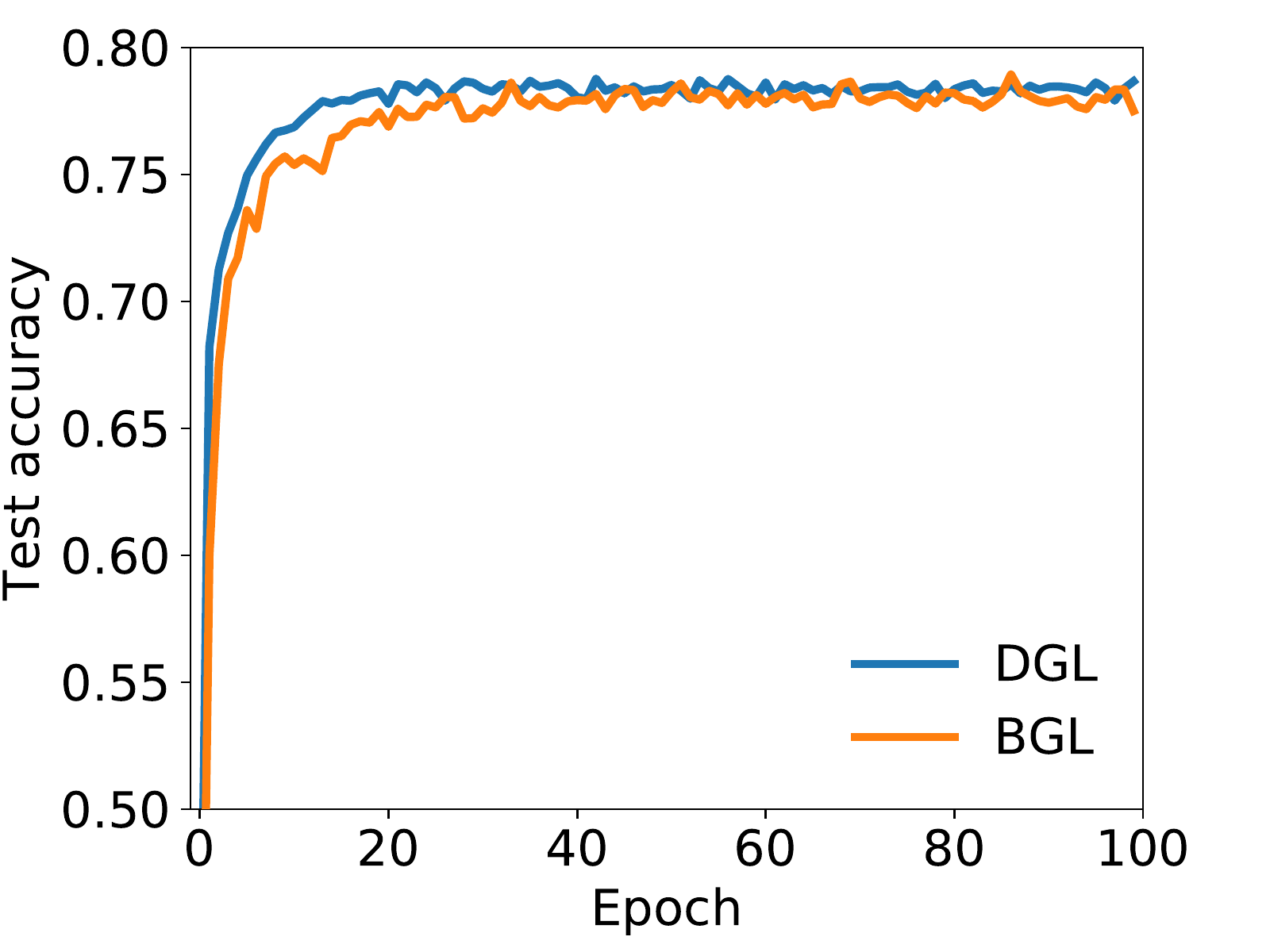}
			\caption{GraphSAGE}
		\end{subfigure}
	    \hspace{3mm}
		\begin{subfigure}{0.45\linewidth}
			\centering
			\includegraphics[width=\linewidth]{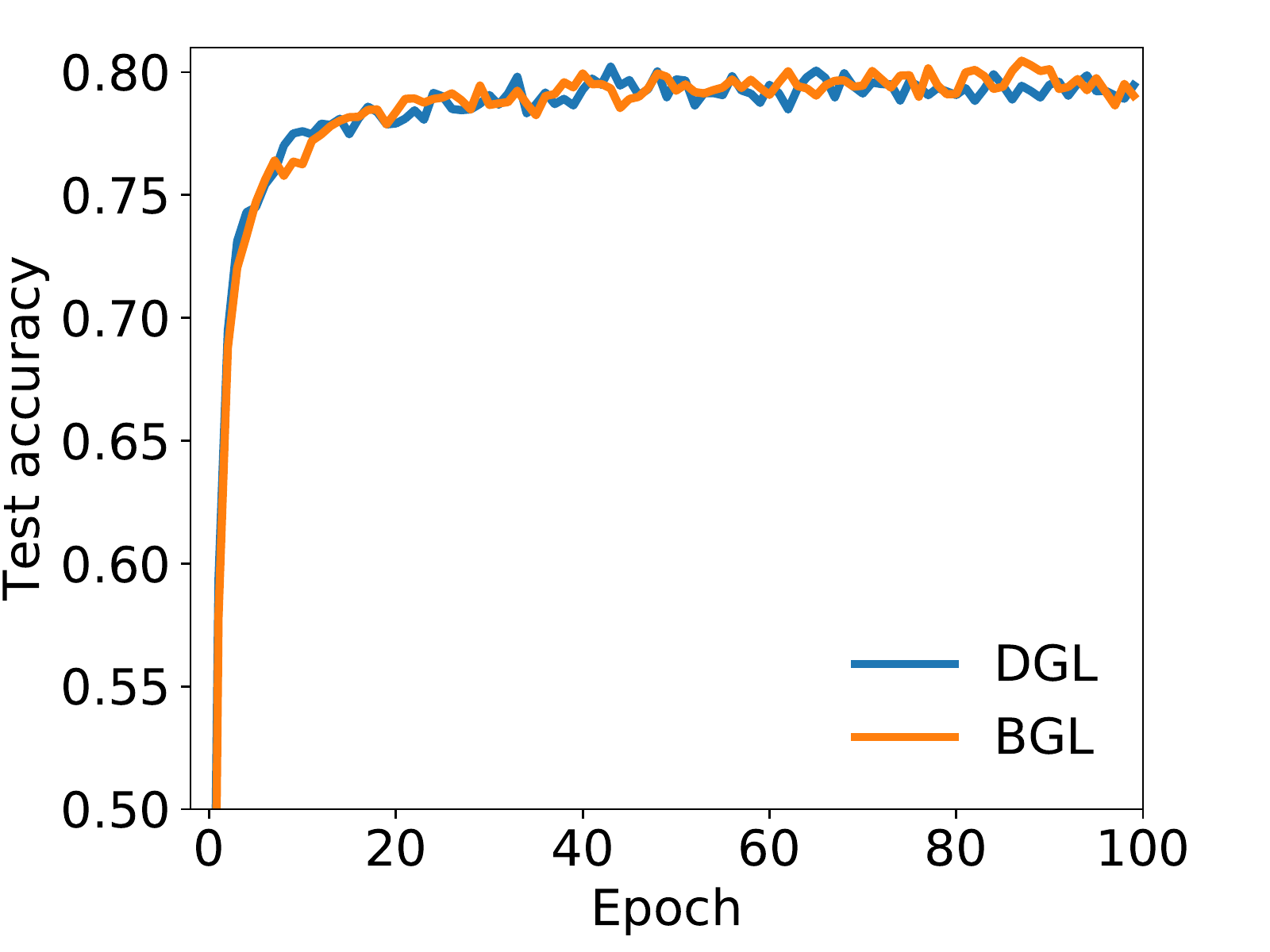}
			\caption{GAT}
		\end{subfigure}
		\vspace{-3mm}
		\caption{Using proximity-aware ordering, {\sysname} converges to the same test accuracy as DGL on Ogbn-products.}
		\label{fig:graphsage_accuracy}
\end{minipage}
\vspace{-5mm}
\end{figure}

To verify the correctness of {\sysname}, %using proximity-aware ordering, 
we evaluate the test accuracy on GCN, GAT and GraphSAGE with %datasets 
Ogbn-products and Ogbn-papers.
We run 100 epochs for each training task.
DGL uses random ordering while {\sysname} uses proximity-aware ordering.
As shown in Table \ref{tab:model_accuracy} and Figure \ref{fig:graphsage_accuracy}, {\sysname} converges to almost the same accuracy as the original DGL. 

\begin{table}[!t]
    \centering
 \begin{minipage}[]{\columnwidth}
        \centering
        \small
        \caption{Test accuracy on different models and graphs. {\sysname} achieves the same test accuracy as DGL.}
        \label{tab:model_accuracy}
        \begin{tabular}{c|c|l|l|c}
        \toprule
        \textbf{Graph} & \textbf{System} & \textbf{GCN} & \textbf{GAT}     & \textbf{GraphSAGE}\\
        \midrule
        Ogbn- & DGL  & 0.775 & 0.792 & 0.789 \\
        %\hline
        products& {\sysname} & 0.773 & 0.796 & 0.789 \\
        \midrule
        Ogbn- & DGL  & 0.460 & 0.517 & 0.460 \\
        %\hline
        papers& {\sysname} & 0.461 & 0.516 & 0.458 \\
        \bottomrule
        \end{tabular}
    \end{minipage}
\vspace{-3mm}
   \end{table}
   
\section{Related Work}
\label{sec:related}

\mypara{Graph Partition Algorithms.} 
Graph partitioning is widely adopted when processing large graphs. 
NeuGraph~\cite{DBLP:conf/usenix/MaYMXWZD19} leverages the Kernighan-Lin~\cite{DBLP:journals/bstj/KernighanL70} algorithm to partition graphs into chunks with different sparsity levels. 
Cluster-GCN~\cite{DBLP:conf/kdd/ChiangLSLBH19} constructs the training batches based on the METIS~\cite{george1998metis} algorithm, together with a stochastic multi-clustering framework to improve model convergence. 
When dealing with large graphs in distributed GNN training, partition algorithms, such as Random, Round-Robin and Linear Deterministic Greedy~\cite{DBLP:journals/pvldb/AbbasKCV18}, are often used~\cite{ali2019euler, DBLP:journals/corr/abs-1909-01315, DBLP:journals/pvldb/ZhuZYLZALZ19, DBLP:journals/pvldb/ZhangHL0HSGWZ020}. They incur low partitioning overhead while not ensuring partition locality. 

\mypara{GNN Training Frameworks.}
In recent years, new specialized frameworks have been proposed upon existing deep learning frameworks to provide convenient and efficient graph operation primitives for GNN training~\cite{DBLP:journals/corr/abs-1909-01315, DBLP:journals/corr/abs-1903-02428, ali2019euler, DBLP:conf/usenix/MaYMXWZD19, DBLP:journals/pvldb/ZhuZYLZALZ19}. Other than  DGL~\cite{DBLP:journals/corr/abs-1909-01315}, Euler~\cite{ali2019euler} and PyG~\cite{DBLP:journals/corr/abs-1903-02428}, NeuGraph~\cite{DBLP:conf/usenix/MaYMXWZD19}  translates graph-aware computation on dataflow and recasts graph optimizations to support parallel computation for GNN training. But 
it can only train small graphs on multi-GPUs in a single machine. 
AliGraph~\cite{DBLP:journals/pvldb/ZhuZYLZALZ19} is a GNN system that consists of distributed graph storage, optimized sampling operators and runtime to support both existing GNNs and in-house developed ones for different scenarios. AGL~\cite{DBLP:journals/pvldb/ZhangHL0HSGWZ020} is a scalable and integrated GNN system, implemented on MapReduce~\cite{dean2008mapreduce} that guarantees good system properties. However, neither Aligraph nor AGL exploits GPU acceleration.

\mypara{GNN Training Acceleration.}
Various systems have been devoted to improving GNN training performance.

Some works~\cite{DBLP:conf/mlsys/JiaLGZA20,DBLP:conf/usenix/MaYMXWZD19,thorpe2021dorylus,wang2021gnnadvisor,cai2021dgcl} target full-batch training.
GNNAdvisor~\cite{wang2021gnnadvisor} explores the GNN input properties and proposes a 2D workload management and specialized memory customization for system optimizations. 
DGCL~\cite{cai2021dgcl}  proposes a communication planning algorithm to optimize GNN communication among multiple GPUs with METIS partition. Both projects assume that graphs are stored in a single machine. 

Some works~\cite{lin2020pagraph,gandhi2021p3,DBLP:journals/pvldb/ZhuZYLZALZ19} target mini-batch training. PaGraph~\cite{lin2020pagraph} adopts static GPU caching for nodes with high degrees. Still, it assumes that graph can be loaded in a single machine, hence infeasible for large graphs.
\new{$P^3$~\cite{gandhi2021p3} reduces retrieving feature traffic by combining model parallelism and data parallelism. 
However, hybrid parallelism in $P^3$ incurs extra synchronization overhead. Its performance suffers when hidden dimensions exceeds 128 (a common practice in modern GNNs).
Further, $P^3$ overlooked the subgraph sampling stage. In subgraph sampling, random hashing partitioning leads to extensive cross-partition communication.}

Some works try to improve graph sampling performance on GPUs, such as NextDoor~\cite{DBLP:conf/eurosys/JangdaPGS21} and C-SAW~\cite{DBLP:conf/sc/PandeyLHLL20}.
However, their performance is limited by small GPU memory.
Hence, they are not suitable for giant graphs.

\vspace{-0.2in}
\section{Conclusion}
\label{sec:conclu}

We present {\sysname}, a GPU-efficient GNN training system for large graph learning that focuses on removing the data I/O and preprocessing bottleneck, to achieve high GPU utilization and accelerate training.
To minimize feature retrieving traffic, we propose a dynamic feature cache engine with proximity-aware ordering, and find a sweet spot of low overhead and high cache hit ratio.
{\sysname} employs a novel graph partition algorithm tailored for sampling algorithms to minimize cross-partition communication during sampling.   
We further optimize the resource allocation of data preprocessing 
using profiling-based resource isolation. Our extensive experiments demonstrate that {\sysname} significantly outperforms existing GNN training systems by 20.68x on average. 
We will open-source {\sysname} in the future and hope to continue evolving it with the community.

%\begin{acks}
%To Robert, for the bagels and explaining CMYK and color spaces.
%\end{acks}

\bibliographystyle{plain}
\bibliography{reference}

%\appendix

\end{document}